%% file: main.tex
\documentclass[runningheads]{llncs}
\usepackage{eccv}
\usepackage{graphicx}
\usepackage{booktabs}
\usepackage{amsmath}
\usepackage{amssymb}
\usepackage{microtype}
\usepackage{placeins}
\usepackage[accsupp]{axessibility}
\usepackage[breaklinks,colorlinks,citecolor=eccvblue]{hyperref}
\newcommand{\citep}[1]{\cite{#1}}

\newcommand{\drdq}{DRDQ-32B}
\newcommand{\scode}{$S_{\text{code}}$}
\newcommand{\svisual}{$S_{\text{visual}}$}
\newcommand{\sphys}{$S_{\text{phys}}$}
\AtEndEnvironment{abstract}{}
\hypersetup{
 pdftitle={PhysCodeBench: Benchmarking Physics-Aware Symbolic Simulation of 3D Scenes via Self-Corrective Multi-Agent Refinement},
 pdfauthor={Tianyidan Xie, Peiyu Wang, Jiaxin Hu, Yuyi Qian, Yuxuan Wang, Shenyi Wang, Rui Ma, Yanlun Peng, Lanjun Wang, Ying Tai, Jian Yang, Zili Yi},
 pdfsubject={Physics simulation code generation, PhysCodeEval, and multi-agent refinement},
 pdfkeywords={PhysCodeBench, physics simulation, code generation, multi-agent systems}
}
\begin{document}
\title{PhysCodeBench: Benchmarking Physics-Aware Symbolic Simulation of 3D Scenes via Self-Corrective Multi-Agent Refinement}
\titlerunning{PhysCodeBench}
\author{Tianyidan~Xie\inst{1}\fnmsep\thanks{Corresponding authors: Tianyidan Xie (\texttt{sealical@outlook.com}) and Zili Yi (\texttt{yi@nju.edu.cn}).} \and
Peiyu~Wang\inst{2} \and
Jiaxin~Hu\inst{1} \and
Yuyi~Qian\inst{1} \and\\
Yuxuan~Wang\inst{1} \and
Shenyi~Wang\inst{1} \and
Rui~Ma\inst{3} \and
Yanlun~Peng\inst{4} \and\\
Lanjun~Wang\inst{5} \and
Ying~Tai\inst{1} \and
Jian~Yang\inst{1} \and
Zili~Yi\inst{1,\star}}
\authorrunning{T. Xie et al.}
\institute{Nanjing University \and
Skywork AI \and
Jilin University \and
Great Wall Motor \and
Tianjin University}
\maketitle
\input{main_body}
\FloatBarrier
\bibliographystyle{splncs04}
\bibliography{physcodebench}
\clearpage
\appendix
\section*{Technical Appendix}
\input{appendix}
\end{document}

%% file: main_body.tex
\begin{abstract}
Translating natural-language descriptions of physical phenomena into executable simulation code requires both programming expertise and physical reasoning. Current large language models (LLMs) lack this combination: they frequently produce code that runs but simulates the wrong physics. We introduce PhysCodeBench, the first benchmark for this task, with 1,200 expert-validated examples spanning four physical domains. Its evaluation suite, PhysCodeEval, goes beyond executability and visual fidelity to measure physical correctness directly from the engine state via conservation-law residuals and expert-written assertions, and supports cross-engine evaluation to disentangle physics reasoning from API fluency. As a reference method, we propose the Self-Corrective Multi-Agent Refinement Framework (SMRF), which decouples physics-aware error correction from code generation through specialized agents. This design is motivated by our finding that targeted correction, rather than generic iterative refinement, is the key driver of physical accuracy. SMRF nearly triples the physical-assertion pass rate of the best proprietary baseline (70.6\% vs.\ 23.8\%) and retains its advantage under cross-engine transfer.

\end{abstract}

\section{Introduction}

Encoding physical laws in executable simulation code is central to robotics, embodied AI, and scientific computing, yet it remains a formidable challenge that demands both programming expertise and physical reasoning. Whether large language models (LLMs) can bridge this gap is an open question, but answering it requires a benchmark in which physical understanding is \emph{observable} in the code and the trajectories it produces.

By symbolic simulation we mean writing a program that instantiates a physical 3D scene in a simulator, so that the physics is carried by the code rather than by a learned generative model. The task compounds three difficulties: the code must execute without runtime errors, it must implement the intended physical laws and parameters, and it must respect boundary conditions and numerical-stability constraints.

\begin{figure}[t]
    \centering
    \includegraphics[width=0.99\textwidth]{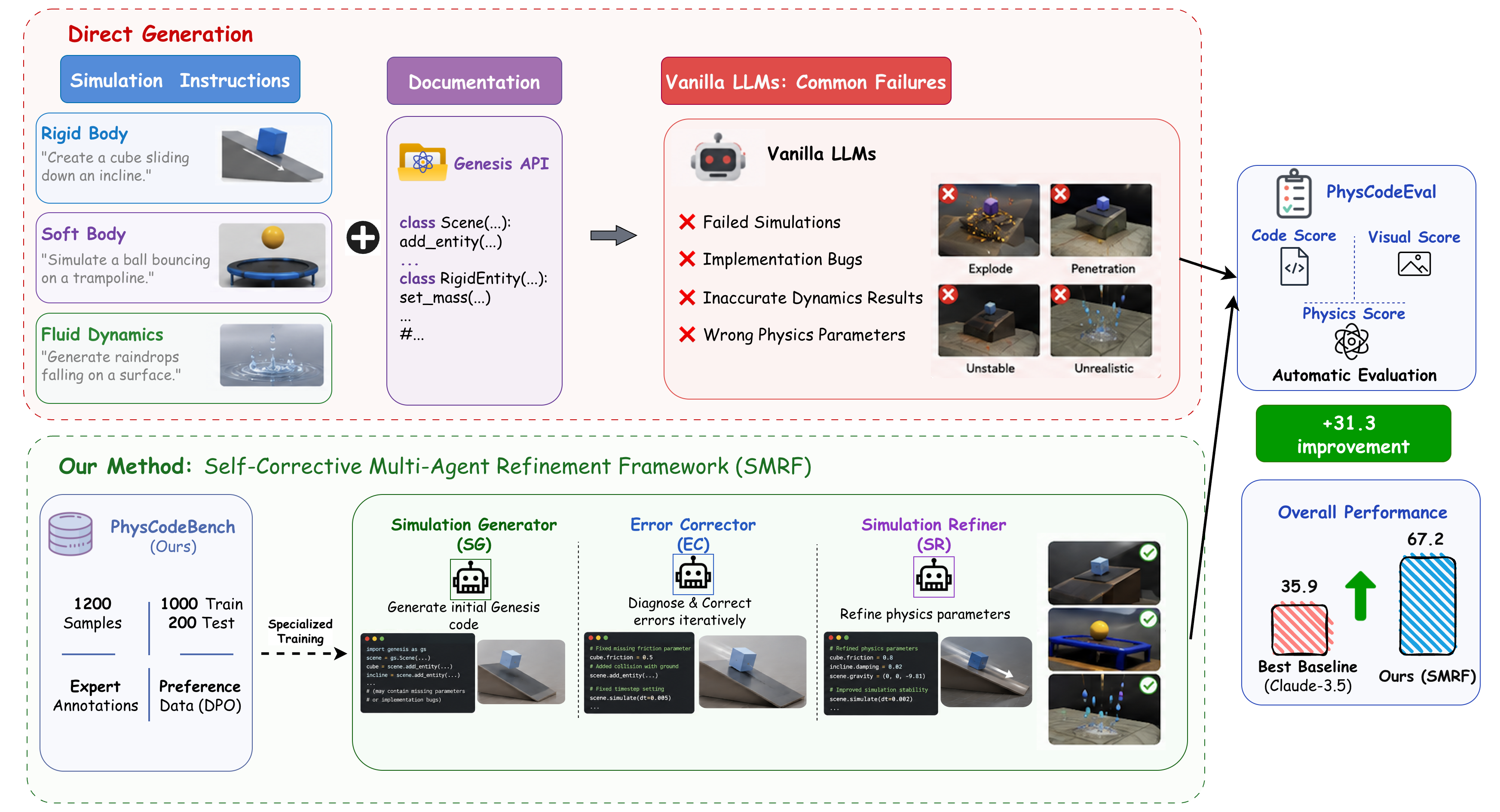}
    \caption{Overview of PhysCodeBench and SMRF. Top: vanilla LLMs prompted with a simulation instruction and engine documentation produce failed, unstable, or physically incorrect simulations. Bottom: the SMRF reference framework decomposes the task across specialized generation, error-correction, and refinement agents. All outputs are scored by PhysCodeEval across code-level, visual, and direct physical-correctness axes.}
    \label{fig:teaser}
\end{figure}

Consider seemingly simple tasks like creating a cube sliding down an incline, simulating a ball bouncing on a trampoline, or generating raindrops falling on a surface. These scenarios require precise implementation of gravity, elasticity, surface tension, and collision dynamics. As illustrated in Figure~\ref{fig:teaser}, vanilla LLMs struggle with these tasks, producing failed simulations, implementation bugs, inaccurate dynamics, and incorrect physics parameters. Even when the code executes successfully, GPT-4o~\citep{hurst2024gpt} and Claude-3.5-Sonnet~\citep{claude_35} frequently generate simulations whose object interactions deviate from the described scenario: execution is only the first hurdle.

Existing code-generation benchmarks~\citep{chen2021evaluating,roziere2023code} lack the evaluation criteria this task demands. Visual plausibility alone is insufficient: a smooth video can hide a violated conservation law. What is needed is \emph{direct} measurement of physical correctness in the simulator's state space, together with cross-engine evaluation that disentangles physics reasoning from memorized API fluency.

We address this gap with three key contributions:
\begin{itemize}
    \item \textbf{PhysCodeBench}: the first benchmark for physics-aware symbolic simulation of 3D scenes, comprising 1,200 expert-validated examples across rigid-body, soft-body, fluid dynamics, and mechanics domains, with seed-level train/test isolation and a three-fold contamination audit.
    \item \textbf{PhysCodeEval}: an evaluation suite that complements execution and visual metrics with direct physical-correctness measurements read from the engine state, including conservation-law residuals and expert-written physical assertions, and supports cross-engine evaluation on a MuJoCo-native subset.
    \item \textbf{SMRF}: a reference multi-agent framework that decouples physics-aware error correction from code generation, nearly tripling the assertion pass rate of the best proprietary baseline and retaining its advantage across engines.
\end{itemize}

\section{Related Work}

\paragraph{Physics simulation and symbolic code generation.}
Prior work studies physical reasoning through video prediction~\citep{bear2021physion,tung2023physion++,yi2019clevrer,zheng2024contphy} or domain-specific solvers~\citep{takamoto2022pdebench,qiu2025phybench}, but none targets executable simulation code. LLM-based approaches invoke engines for reasoning~\citep{liu2022mind,cherian2024llmphy} or generate control programs, reward functions, and scene or task specifications~\citep{code_as_policies,huang2023voxposercomposable3dvalue,simgen,wang2024gensim,wang2024robogen,hu2024scenecraft,ma2024eureka}, yet none requires authoring a complete physics simulation whose physical correctness is then measured.

\paragraph{Code generation and its evaluation.}
Code generation has advanced with models like Codex~\citep{chen2021evaluating}, CodeLlama~\citep{roziere2023code}, DeepSeek-R1~\citep{guo2025deepseek}, and Qwen-2.5~\citep{yang2024qwen2}. These are evaluated on general programming benchmarks such as HumanEval~\citep{chen2021evaluating} and MBPP~\citep{austin2021program}, where correctness is fully captured by input--output tests. Physics simulation breaks this assumption: code can pass every syntactic check yet simulate the wrong world. Fine-tuning approaches like WizardCoder~\citep{luo2023wizardcoder} show specialization improves domain performance, motivating our multi-agent approach, but evaluation methodology for \emph{physically} correct code has been missing. PhysCodeEval fills this gap with state-space checks.

\paragraph{Multi-agent systems and preference alignment.}
Multi-agent frameworks like AutoGen~\citep{wu2023autogen} and ChatDev~\citep{qian2023chatdev} demonstrate how specialized agents collaborate on complex tasks. In code generation, approaches like Self-Debugging~\citep{chen2023teaching} and CodeChain~\citep{le2023codechain} show iterative refinement improves quality, but lack physics-specific knowledge. We show that generic self-refinement leaves a 26-point gap to specialized correction. Preference alignment (RLHF~\citep{ouyang2022training}, DPO~\citep{rafailov2023direct}) improves LLM quality generally, and we apply it to physics-specific agent roles with expert simulation preferences.

\section{The PhysCodeBench Benchmark}

\subsection{Overview}

PhysCodeBench comprises 1,200 examples spanning rigid-body physics (410), soft-body physics (310), fluid dynamics (275), and mechanics (205), with two difficulty levels (709 easy / 491 hard). Hard examples involve three or more interacting physical laws, multi-phase scenarios, or parameter regimes requiring precise tuning. Each example pairs a natural-language instruction with an expert-validated reference implementation for the Genesis physics engine~\citep{Genesis}, plus difficulty, physical-law, object-type, and preference metadata. The dataset is split into 1,000 training and 200 testing examples with balanced domain coverage (test: 68 rigid-body, 52 soft-body, 46 fluid, 34 mechanics) and 116 easy / 84 hard examples. Dataset examples and complete statistics are provided in the technical appendix.

\subsection{Dataset Curation Process}

\begin{figure}[t]
    \centering
    \includegraphics[width=0.99\textwidth]{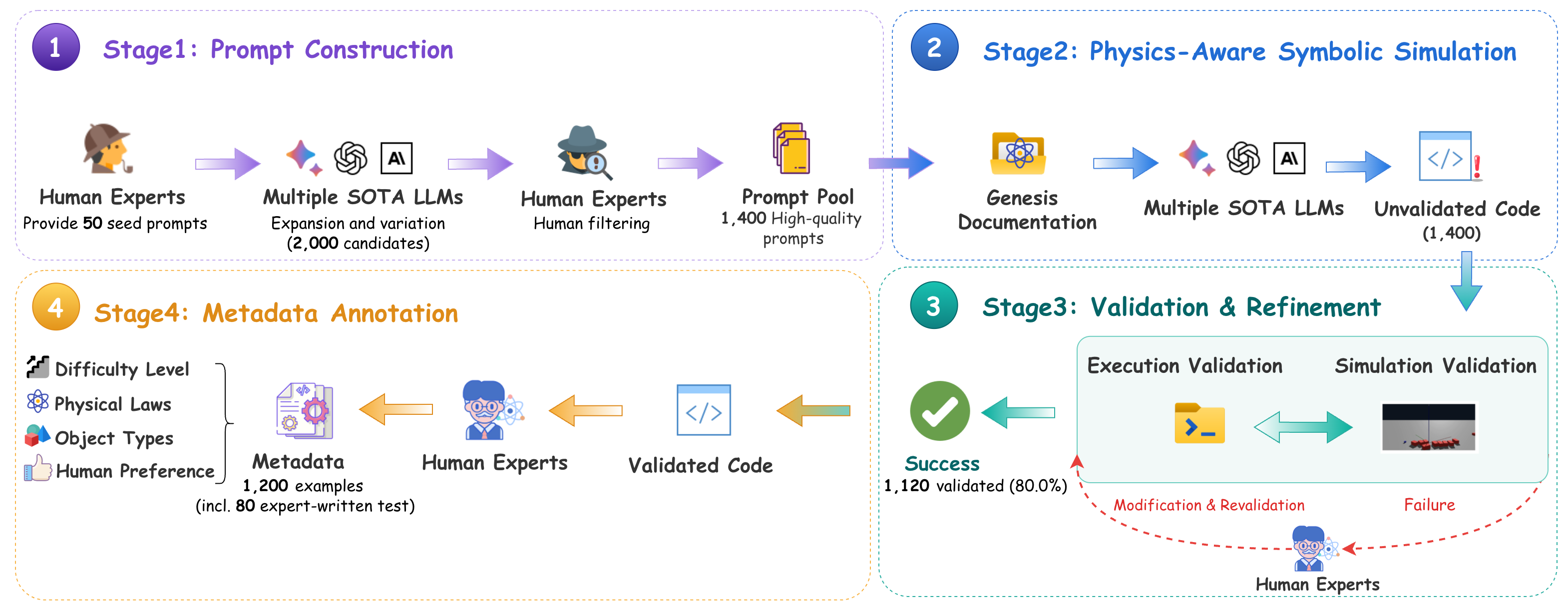}
    \caption{PhysCodeBench dataset curation pipeline. Stage 1: 50 expert seed prompts are expanded by multiple LLMs into 2,000 candidate instructions, and human filtering retains a pool of 1,400. Stage 2 (physics-aware symbolic simulation): multiple LLMs generate implementations against the Genesis documentation. Stage 3: two-stage validation (execution, then simulation faithfulness) with an expert modification loop. In total, 430 candidates (30.7\%) pass both stages on the first attempt and 1,120 (80.0\%) are retained after at most three expert revision rounds. Together with 80 expert-authored test instructions, this yields the final 1,200 examples. Stage 4: experts annotate difficulty, physical laws, object types, and preferences.}
    \label{fig:data_pipe}
\end{figure}

Construction follows a four-stage pipeline (Figure~\ref{fig:data_pipe}).

\paragraph{Stage 1: Prompt construction.}
Human experts first draft 50 seed prompts arranged in a $4\times2$ domain-by-difficulty design covering 32 distinct combinations of physical laws. These seed prompts are then provided to multiple state-of-the-art proprietary models~\citep{hurst2024gpt,claude_35,gemini_2_pro,github_copilot} for expansion and variation. Expansion is constrained to stay within a single seed's scenario family, forbidding cross-seed template merging. Human experts subsequently filter the machine-generated candidates for clarity, physical well-posedness, and diversity to form the Prompt Pool. Through this process, we generated 2,000 instruction candidates from the 50 seed prompts and retained a pool of 1,400.

\paragraph{Stage 2: Physics-aware symbolic simulation.}
For each prompt in the Prompt Pool, we generate simulation code with the same set of proprietary models, each given the prompt, the Genesis documentation, and the formatting requirements. Using multiple models keeps the reference implementations from inheriting a single model's coding style.

\paragraph{Stage 3: Code validation and refinement.}
The generated code undergoes a two-stage validation process: execution validation and simulation validation. Execution validation tests whether the code executes without errors in the Genesis environment. Simulation validation examines whether the executed code produces simulation visualizations that match the intended physical scenario. Only 430 of the 1,400 pooled candidates (30.7\%) pass both validations on the first attempt. When validation fails, human experts manually modify the code and repeat the validation process, with up to three modification attempts allowed. In total, 1,120 pooled instructions (80.0\%) yield a validated implementation and 280 are discarded. All retained code is additionally rewritten by experts (mean CodeBLEU~\citep{ren2020codebleu} 0.46 against the original LLM draft), so reference solutions are not verbatim LLM outputs. Adding 80 test instructions authored, together with their reference implementations, entirely by experts (see Test-Set Integrity) yields the final 1,200 examples.

\paragraph{Stage 4: Metadata annotation.}
After validating the prompt--code pairs, we enrich each example with comprehensive metadata. Three annotators label difficulty (mean pairwise Cohen's $\kappa=0.78$) and physical-law tags (mean pairwise Jaccard-based $\kappa=0.71$). To prepare preference data for training, we generate two different validated implementations for 600 scenarios. Five experts provide 1,980 pairwise preference judgments over these 600 implementation pairs (3.3 raters per pair, Krippendorff's $\alpha=0.68$), and the majority label per pair yields the preference data used for training.

\subsection{Test-Set Integrity}
\label{sec:integrity}

Because both the candidate instructions and draft code originate partly from proprietary LLMs, we protect the test split by construction and audit it post hoc.

\paragraph{Seed isolation.} 15 of the 50 seed prompts are reserved for the test split: their expansions never enter training data. Of the 1,120 pool-derived examples, 1,000 (from the 35 training seeds) form the training set and 120 (from the 15 held-out seeds) enter the test set. The remaining 80 test instructions are newly authored by experts with no LLM in the loop. All Error-Corrector tetrads and DPO preference scenarios are drawn exclusively from the 1,000 training instructions. No artifact derived from a test instruction appears in any agent's training data.

\paragraph{Contamination audit.} We measure train--test proximity three ways: (i) \emph{semantic}: embedding cosine similarity between each test instruction and its nearest training neighbor has mean 0.41 and max 0.63, and zero pairs exceed the 0.85 near-duplicate threshold; (ii) \emph{lexical}: BM25 retrieval~\citep{robertson2009bm25} gives a maximum token-level Jaccard overlap of 0.27; (iii) \emph{code-level}: zero-shot samples (five per prompt) from the strongest baselines never exceed CodeBLEU 0.7 against the corresponding test reference implementations, giving no evidence that test solutions are memorized. Note that pretraining contamination, if present, would inflate the \emph{proprietary baselines} rather than our open-weight reference method, making the reported gaps conservative.

\section{The PhysCodeEval Evaluation Suite}
\label{sec:eval}

PhysCodeEval scores a generated simulation on three axes. Its first two form the 100-point score $\text{Total} = S_{\text{code}} + S_{\text{visual}}$ used throughout our experiments. The third axis measures physical correctness directly and is reported as a first-class metric alongside the total.

\paragraph{Code quality score \scode{} (50 pts).}
\scode{} evaluates whether the generated code executes successfully ($S_{\text{exec}}$, 25 points, awarded in full or not at all) and produces the expected simulation artifact ($S_{\text{file}}$, 25 points, scored by a fixed partial-credit rubric over existence, size, and format compliance).

\paragraph{Simulation fidelity score \svisual{} (50 pts).}
\svisual{} assesses the generated simulation quality through two metrics: $S_{\text{clip}}$ measures semantic alignment between simulation video and instruction using ClipScore~\citep{hessel2021clipscore}, while $S_{\text{motion}}$ evaluates physical realism via a flow-based motion-smoothness measure inspired by VBench~\citep{huang2024vbench}, detecting anomalies such as jitter, unrealistic accelerations, or object intersections. Each component contributes up to 25 points. These perceptual metrics are useful but insufficient alone: a visually smooth video can violate conservation laws. This motivates the third axis.

\paragraph{Direct physical-correctness measurements.}
Both measurements below are computed from the \emph{engine state interface} rather than from rendered video, eliminating vision-pipeline error.
\begin{itemize}
    \item \emph{Assertion score \sphys{}.} During annotation, experts wrote 1--3 executable physical assertions per test instruction (436 in total), e.g., ``first rebound height $\in[0.6h_0, 0.95h_0]$'' or ``$\geq$3 concentric ripples within 0.3\,s of impact.'' Assertions are authored from the instruction text alone, blind to any implementation, and each is reviewed by a second expert. The expert reference implementations pass 98.6\% of them. \sphys{} is the fraction of assertions passed, where every assertion of a program that fails to execute counts as failed. Because assertion counts vary per scenario, \sphys{} is reported alongside the 100-point total rather than folded into it.
    \item \emph{Conservation checks (companion diagnostic).} For every step we read center-of-mass positions, velocities, and angular velocities. For scenario labels with well-defined invariants we compute residuals: momentum drift $\lVert\Delta p\rVert/\lVert p_0\rVert$ for collision scenarios, fitted acceleration versus $g$ for free fall, and kinetic-energy retention for elastic versus dissipation bounds for inelastic contact, each passed at a 5\% tolerance and reported on the subset where invariants are defined. Unlike \sphys{}, which is a score and so counts every assertion of a non-executing program as failed, these residuals are a diagnostic evaluated only on runs that produced a trajectory, which is what lets them separate physical error from execution failure. For dissipative scenarios (soft bodies, fluids) where global conservation is ill-defined, we check scenario-appropriate necessary conditions instead: monotone energy dissipation, terminal-state stability, and non-interpenetration.
\end{itemize}

\paragraph{Cross-engine protocol.}
To separate physics reasoning from single-engine API fluency, 120 test instructions carry MuJoCo-native reference implementations and assertion sets, enabling the code, perceptual, and assertion axes to be computed on MuJoCo~\citep{mujoco}. This makes cross-engine evaluation a property of the \emph{benchmark}, independent of any particular method's adaptation strategy.

\section{Methodology}
% textwidth, columnwidth
\begin{figure}[t]
    \centering
    \includegraphics[width=0.85\textwidth]{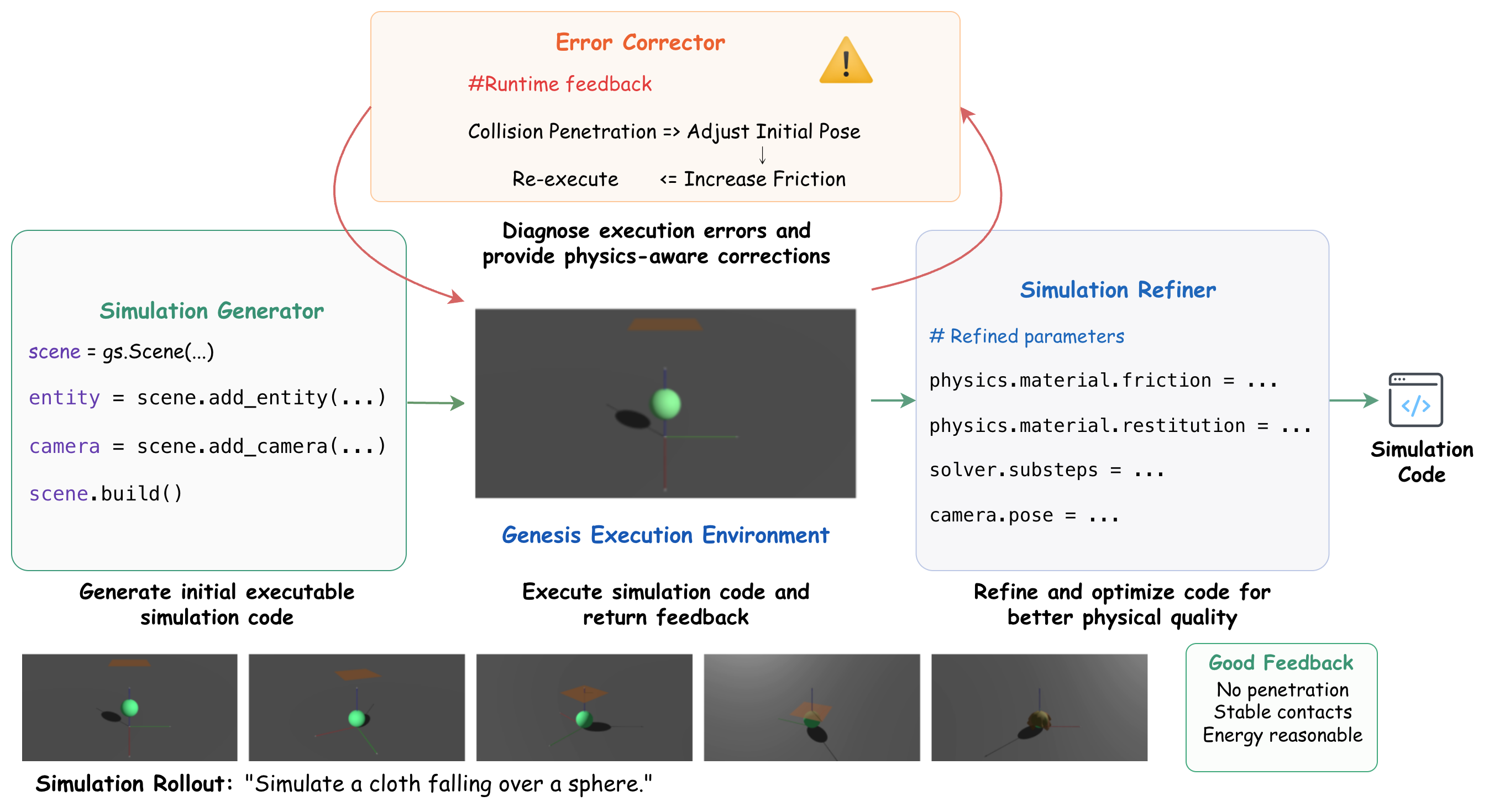}
    \caption{The Self-Corrective Multi-Agent Refinement Framework (SMRF). The Simulation Generator produces initial code, which is executed in the Genesis environment. Runtime feedback drives the Error Corrector, which diagnoses execution errors and provides physics-aware corrections (up to three rounds). Once execution succeeds, the Simulation Refiner tunes physical parameters and code quality under preference alignment. Bottom: a simulation rollout for ``Simulate a cloth falling over a sphere.''}
    \label{fig:smrf}
\end{figure}

The Self-Corrective Multi-Agent Refinement Framework (SMRF) splits generation, correction, and refinement across specialized agents (Figure~\ref{fig:smrf}). All agents are initialized from DeepSeek-R1-Distill-Qwen-32B (\drdq{})~\citep{drdq_32b} and specialized through role-specific training.

\paragraph{Agents and control flow.}
The \textbf{Simulation Generator (SG)} is trained on PhysCodeBench's 1,000 training pairs using supervised fine-tuning (SFT) and produces initial code based on the instruction prompt, with the standard objective
\begin{equation}
    \mathcal{L}_{\text{SFT}} = -\textstyle\sum_{i=1}^{N} \log p_\theta(y_i \mid x_i),
\end{equation}
where $x_i$ represents the instruction prompt and $y_i$ the reference implementation. The generated code is executed in the Genesis environment to identify runtime errors or execution failures. When execution fails, the \textbf{Error Corrector (EC)} diagnoses specific errors and proposes corrections based on its training on 520 tetrads $(x, c_{\text{bug}}, e, c_{\text{fix}})$ of (instruction, incorrect code, error description, corrected code), harvested from the validation stage of dataset curation and restricted to training instructions (43\% API misuse, 28\% parameter miscalibration, 18\% boundary conditions, 7\% temporal discretization, 4\% other), with loss computed only on $c_{\text{fix}}$ tokens. The EC attempts to correct errors up to 3 times. If errors persist after these attempts, the framework returns a failure. When execution succeeds, the \textbf{Simulation Refiner (SR)} further optimizes the code by refining physical parameters while maintaining code structure. The SR is first trained with SFT, then aligned with human preferences on the 600 majority-labeled preference pairs (1,980 judgments) through direct preference optimization (DPO)~\citep{rafailov2023direct}:
\begin{equation}
    \mathcal{L}_{\text{DPO}} = -\mathbb{E}\!\left[\log \sigma\!\left(\beta \log \tfrac{p_\theta(y_w|x)}{p_{\text{ref}}(y_w|x)} - \beta \log \tfrac{p_\theta(y_l|x)}{p_{\text{ref}}(y_l|x)}\right)\right],
\end{equation}
where $y_w$ and $y_l$ are the preferred and less preferred implementations and $\beta=0.1$.

\section{Experiments}

\subsection{Experimental Setup}
\label{sec:setup}

\paragraph{Models and baselines.}
We compare our approach against several strong baselines. For proprietary models, we evaluate GPT-4o~\citep{hurst2024gpt}, Claude-3.5-Sonnet~\citep{claude_35}, and Gemini-2.0-Pro~\citep{gemini_2_pro}. From the open-source community, we test DeepSeek-R1~\citep{guo2025deepseek}, \drdq{}~\citep{drdq_32b}, Qwen-2.5-32B~\citep{yang2024qwen2}, and QwQ-32B~\citep{qwq}, spanning general-purpose and reasoning-oriented open models. We also create fine-tuned single-agent variants of \drdq{} using SFT and DPO, which operate without the multi-agent framework. Our SMRF implementation includes three variants: SMRF (base) with unspecialized agents, SMRF + SFT, and SMRF + SFT + DPO (our complete framework).

\paragraph{Training and inference.}
All agents are specialized through role-specific LoRA fine-tuning on 2 NVIDIA A100 80GB GPUs, using a learning rate of $10^{-5}$, batch size of 2, 5 epochs for SFT and 3 epochs for DPO, requiring approximately 34 GPU-hours for the main configuration. During inference, each model is provided with Genesis physics engine documentation: models with long context windows receive the full $\approx$100K-token corpus, while 32K-context models receive $\approx$20K tokens of BM25-retrieved task-relevant sections (a context ablation justifying this protocol is in the technical appendix). Each model generates responses with a maximum length of 4,096 tokens and temperature of 0.1. We perform 5 inference passes per test prompt and report the average score (avg@5). Complete configurations are in the technical appendix.

\subsection{Quantitative Analysis}

\begin{table}[t]
\centering
\small
\renewcommand{\arraystretch}{0.87}
\setlength{\tabcolsep}{4pt}
\begin{tabular}{lcccc}
\toprule
\textbf{Model} & \textbf{Code} & \textbf{Visual} & \textbf{Total} & \textbf{\sphys{}\,(\%)} \\
\midrule
\multicolumn{5}{l}{\textit{Proprietary (zero-shot)}} \\
GPT-4o             & 15.7 & 18.1 & 33.8 & 21.6 \\
Claude-3.5-Sonnet  & 17.0 & 18.9 & \underline{35.9} & \underline{23.8} \\
Gemini-2.0-Pro     & 14.8 & 16.8 & 31.6 & 19.7 \\
\midrule
\multicolumn{5}{l}{\textit{Open-source (zero-shot)}} \\
DeepSeek-R1        & 13.8 & 15.7 & 29.5 & 17.9 \\
\drdq{}            & 12.0 & 15.6 & 27.6 & 16.2 \\
Qwen-2.5-32B       &  0.8 &  1.1 &  1.9 &  0.9 \\
QwQ-32B            &  6.7 &  8.8 & 15.5 &  7.4 \\
\midrule
\multicolumn{5}{l}{\textit{Single-agent fine-tuned}} \\
\drdq{} + SFT       & 17.2 & 18.2 & 35.4 & 28.9 \\
\drdq{} + SFT + DPO & 18.4 & 19.1 & 37.5 & 32.1 \\
\midrule
\multicolumn{5}{l}{\textit{Multi-agent (SMRF)}} \\
Base (vanilla agents) & 16.2 & 17.5 & 33.7 & 30.4 \\
+ SFT                 & 30.5 & 31.1 & 61.6 & 63.5 \\
+ SFT + DPO           & \textbf{33.2} & \textbf{34.0} & \textbf{67.2} & \textbf{70.6} \\
\midrule
Expert reference (oracle) & 50.0 & 44.1 & 94.1 & 98.6 \\
\multicolumn{3}{l}{\textit{vs.\ best proprietary baseline}} & \textbf{+31.3} & \textbf{+46.8} \\
\bottomrule
\end{tabular}
\caption{Results on the PhysCodeBench test set (200 examples, avg@5). Code and Visual are each out of 50, Total out of 100. \sphys{} is the physical-assertion pass rate. The oracle row scores the expert reference implementations and gives the attainable ceiling. Bold = best system, underline = best proprietary baseline. \drdq{} = DeepSeek-R1-Distill-Qwen-32B.}
\label{tab:main}
\end{table}

Table~\ref{tab:main} presents the performance of all approaches on the PhysCodeBench test set. Several key findings emerge from these results. First, the task is far from solved: the best zero-shot model, Claude-3.5-Sonnet, reaches 35.9 out of 100 points and passes only 23.8\% of physical assertions despite full engine documentation in context. Second, direct physical measurement changes the picture. The gap between the strongest fine-tuned single agent and SMRF is 29.7 points on the 100-point total but 38.5 points on \sphys{}. The two axes even reorder systems: \drdq{}+SFT ranks below Claude-3.5-Sonnet on the total (35.4 vs.\ 35.9) yet above it on \sphys{} (28.9\% vs.\ 23.8\%), and SMRF (base) shows the same inversion against GPT-4o (33.7 vs.\ 33.8 on the total, 30.4\% vs.\ 21.6\% on \sphys{}). A leaderboard built only from execution and perceptual scores would therefore rank these systems wrongly on physics, which is precisely what the third axis is for. Qwen-2.5-32B is an outlier at 1.9 points: it rarely emits a complete runnable Genesis script, so execution succeeds on under 5\% of samples and both \scode{} and \sphys{} collapse. This is a property of the model rather than of our harness, since two close relatives score far higher under the identical protocol, namely QwQ-32B at 15.5 and the DeepSeek-R1 distillation of the same backbone at 27.6. Notably, SMRF with vanilla agents nearly doubles \sphys{} over its \drdq{} backbone (16.2\%$\rightarrow$30.4\%) while the total rises only 6.1 points, showing that execution-feedback-driven correction improves physics disproportionately to executability. Third, while SFT alone provides substantial improvements (increasing performance from 33.7 to 61.6 points), adding DPO further enhances performance by an additional 5.6 points by aligning the Simulation Refiner's behavior with expert preferences.

On the conservation subset (102 rigid-body and mechanics test cases), SMRF satisfies momentum- and energy-conservation residuals within the 5\% tolerance in 81.4\% of checks versus 35.3\% for Claude-3.5-Sonnet and 42.2\% for the fine-tuned single agent. Because these residuals are measured only on runs that executed, they isolate the failure mode this benchmark targets: Claude-3.5-Sonnet produces a running simulation that nevertheless violates momentum or energy conservation in roughly two thirds of the cases where the invariant is defined. Per-residual breakdowns are in the technical appendix.

\subsection{Performance Across Difficulty Levels}

\begin{table}[t]
\centering
\small
\renewcommand{\arraystretch}{0.87}
\begin{tabular}{lcc}
\toprule
\textbf{Model} & \textbf{Easy} & \textbf{Hard} \\
\midrule
GPT-4o              & 43.0 & 21.1 \\
Claude-3.5-Sonnet   & 45.2 & 23.1 \\
Gemini-2.0-Pro      & 40.7 & 19.0 \\
\drdq{} + SFT + DPO & 46.9 & 24.5 \\
SMRF (base)         & 42.8 & 21.1 \\
SMRF + SFT          & 72.1 & 47.1 \\
SMRF + SFT + DPO    & \textbf{77.5} & \textbf{53.0} \\
\midrule
Improvement over strongest single agent & +30.6 & +28.5 \\
\bottomrule
\end{tabular}
\caption{Total score by difficulty (116 easy / 84 hard test examples). The improvement row is computed against \drdq{}+SFT+DPO, the strongest non-SMRF system.}
\label{tab:difficulty}
\end{table}

Table~\ref{tab:difficulty} reveals that all models exhibit a substantial performance drop when moving from easy to hard tasks (three or more interacting laws, multi-phase scenarios). Claude-3.5-Sonnet loses 48.9\% of its easy-task score on hard tasks (45.2$\rightarrow$23.1), while SMRF + SFT + DPO loses 31.6\% (77.5$\rightarrow$53.0) and keeps a 28.5-point lead over the strongest single agent, where compound failures of syntax, physics, and stability are most likely.

\begin{figure}[t]
    \centering
    \includegraphics[width=0.99\textwidth]{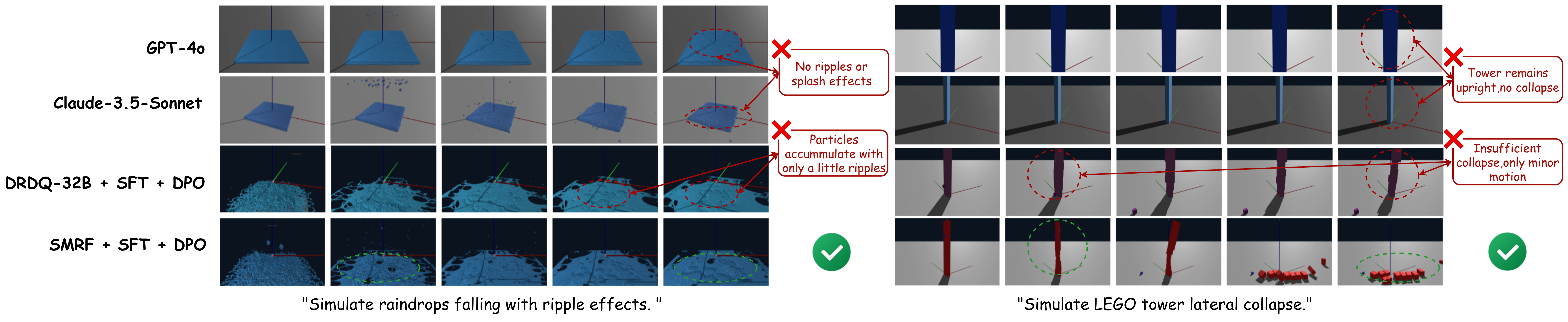}
    \caption{Qualitative comparison. Left (``raindrops falling with ripple effects''): baselines render a nearly static water sheet, while SMRF produces droplets, splashes, and expanding surface waves. Right (``LEGO tower lateral collapse''): baselines leave the tower intact. Only SMRF produces progressive toppling with scattering bricks.}
    \label{fig:qualitative}
\end{figure}

\subsection{Ablation Studies}

\begin{table}[t]
\centering
\small
\renewcommand{\arraystretch}{0.87}
\setlength{\tabcolsep}{4pt}
\begin{tabular}{lcccc}
\toprule
\textbf{Configuration} & \textbf{Code} & \textbf{Visual} & \textbf{Total} & \textbf{\sphys{}\,(\%)} \\
\midrule
Full SMRF + SFT + DPO & 33.2 & 34.0 & \textbf{67.2} & \textbf{70.6} \\
\;--- without EC      & 27.4 & 28.5 & 55.9 & 52.8 \\
\;--- without SR      & 28.8 & 29.6 & 58.4 & 59.8 \\
\;--- SR without DPO  & 30.5 & 31.1 & 61.6 & 63.5 \\
\bottomrule
\end{tabular}
\caption{Ablation study of SMRF components (200 examples, avg@5).}
\label{tab:ablation}
\end{table}

Table~\ref{tab:ablation} reveals the importance of each SMRF component, and every ablation costs more on \sphys{} than on the total. Removing the Error Corrector is the largest single-component effect, at 11.3 points of total and 17.8 points of \sphys{}, since failed executions are never recovered and their assertions all count as failed. Removing the Simulation Refiner costs 8.8 points of total and 10.8 of \sphys{}, reflecting its role in physical-parameter tuning. DPO alignment adds 5.6 points of total and 7.1 of \sphys{} over the SFT-only refiner, which is the SMRF + SFT row of Table~\ref{tab:main} because DPO acts on the SR alone. This is the ablation evidence behind our claim that targeted correction, rather than generic refinement, is what drives physical accuracy.

\subsection{Cross-Engine Evaluation}

\begin{table}[t]
\centering
\small
\renewcommand{\arraystretch}{0.87}
\begin{tabular}{lcc}
\toprule
\textbf{Approach} & \textbf{Adapt.} & \textbf{Total} \\
\midrule
GPT-4o (zero-shot)               & 0   & 26.8 \\
Gemini-2.0-Pro (zero-shot)       & 0   & 24.9 \\
Claude-3.5-Sonnet (zero-shot)    & 0   & 28.5 \\
\drdq{} + SFT + DPO (adapted)    & 250 & 36.7 \\
SMRF (Genesis-trained, transfer) & 0   & 41.2 \\
SMRF + adaptation samples        & 250 & \textbf{58.3} \\
\bottomrule
\end{tabular}
\caption{Cross-engine results on the 120-example MuJoCo-native test subset. ``Adapt.'' = number of MuJoCo API-mapping samples used (0 = zero-shot).}
\label{tab:crossengine}
\end{table}

Two claims must be separated: whether the \emph{benchmark} can evaluate across engines, and whether a particular \emph{method} transfers. The MuJoCo-native subset addresses the first by construction. The proprietary baselines drop about 7 points from their Genesis test-set scores (e.g., Claude-3.5-Sonnet 35.9$\rightarrow$28.5), and their assertion pass rates remain low on both engines (Claude: 20.1\% on MuJoCo). For the second claim, Table~\ref{tab:crossengine} shows that SMRF trained purely on Genesis transfers to MuJoCo at 41.2 points zero-shot, above every zero-shot baseline, and recovers 87.9\% of its Genesis performance on the same 120 instructions (58.3 vs.\ 66.3) with only 250 API-mapping samples. Zero-shot retention is 62.1\%. The EC's correction success rate degrades only from 84\% on Genesis to 79\% under zero-shot transfer: its non-API failure modes (invalid masses, out-of-range coefficients, dimensional inconsistencies) are properties of physics rather than of any engine. PhysCodeBench performance thus reflects transferable physical reasoning, with engine-specific knowledge bridgeable by a small sample budget.

\subsection{User Study and Qualitative Analysis}

\begin{table}[t]
\centering
\small
\renewcommand{\arraystretch}{0.87}
\begin{tabular}{lccc}
\toprule
\textbf{Model} & \textbf{Phys.} & \textbf{Read.} & \textbf{Useful.} \\
\midrule
GPT-4o              & 3.1 & 3.5 & 3.4 \\
Gemini-2.0-Pro      & 3.0 & 3.6 & 3.2 \\
Claude-3.5-Sonnet   & 3.2 & 3.7 & 3.4 \\
\drdq{} + SFT + DPO & 3.4 & 3.5 & 3.6 \\
SMRF + SFT + DPO    & \textbf{4.5} & \textbf{4.2} & \textbf{4.6} \\
\bottomrule
\end{tabular}
\caption{User study results (10 participants, 40 prompts, 1--5 scale): physical accuracy, code readability, overall usefulness.}
\label{tab:userstudy}
\end{table}

We conducted a user study with 10 participants (6 students, 4 professional developers) with physics simulation experience, who rated implementations for 40 prompts on which all compared systems produced executable output. Participants rated each implementation on a 1--5 scale across three dimensions: physical accuracy, code readability, and overall usefulness. SMRF received the highest ratings on all three dimensions (Table~\ref{tab:userstudy}). Inter-rater reliability is ICC(2,1)$=0.74$. To validate our automatic metrics, we computed Spearman correlations with human judgments over 120 paired observations (40 prompts $\times$ 3 systems: Claude-3.5-Sonnet, \drdq{}+SFT+DPO, and SMRF): $S_{\text{clip}}$ correlated with physical accuracy ratings ($\rho=0.82$), and code execution with usefulness ratings ($\rho=0.76$). These correlations are driven largely by differences between systems, so they support the perceptual axes for system-level ranking rather than as substitutes for the state-based measurements. Study protocol details, including recruitment and consent, are in the technical appendix.

Figure~\ref{fig:qualitative} shows the same gap qualitatively: the baselines leave a nearly static fluid surface and an intact tower, while SMRF produces splashes, propagating waves, and progressive toppling with contact-consistent debris. Additional examples with error-correction traces are in the technical appendix.

\subsection{Robustness and Integrity Checks}

\paragraph{Statistical robustness.} Bootstrap 95\% confidence intervals (1,000 resamples) around per-domain totals do not overlap between SMRF and any baseline. Seed-grouped five-fold cross-validation over the 1,120 pool-derived examples (all expansions of a seed share a fold, preserving seed isolation) yields a total-score standard deviation of 1.4 points.

\paragraph{Prompt-style control.} As a stylistic control (no test instruction shares a seed with training data by construction), SMRF's hard-subset score on LLM-expanded held-out-seed instructions (52.1) is within 1.8 points of its score on expert-authored instructions (53.9), so performance does not depend on test prompts sharing the LLM-expansion style of the training data.

\paragraph{Single-agent refinement with equal budget.} Giving GPT-4o, Claude-3.5-Sonnet, and the fine-tuned \drdq{} the same three-round self-correction budget yields at most 41.0 points, 26.2 below SMRF, showing the gain comes from specialized decomposition rather than iteration count (full table, including a CLIP-selected best-of-5 baseline, in the technical appendix).

\section{Conclusion}

We introduce PhysCodeBench, the first benchmark for physics-aware symbolic simulation of 3D scenes: 1,200 expert-validated examples, an evaluation suite that reads conservation residuals and expert assertions directly from engine state, and a MuJoCo-native subset that makes cross-engine evaluation a property of the benchmark itself. Our reference framework SMRF reaches 67.2 points, lifts assertion pass rates from 23.8\% to 70.6\%, and retains 87.9\% of its performance across engines. Limitations remain: the benchmark targets high-level APIs rather than solver implementation, and the error corrector fires only on execution failure. We will release the dataset, SMRF implementation, evaluation framework, and a semi-automated expansion protocol.

%% file: appendix.tex
\section{Dataset Details}

\subsection{Examples and Composition}

\begin{figure}[t]
    \centering
    \includegraphics[width=0.78\textwidth]{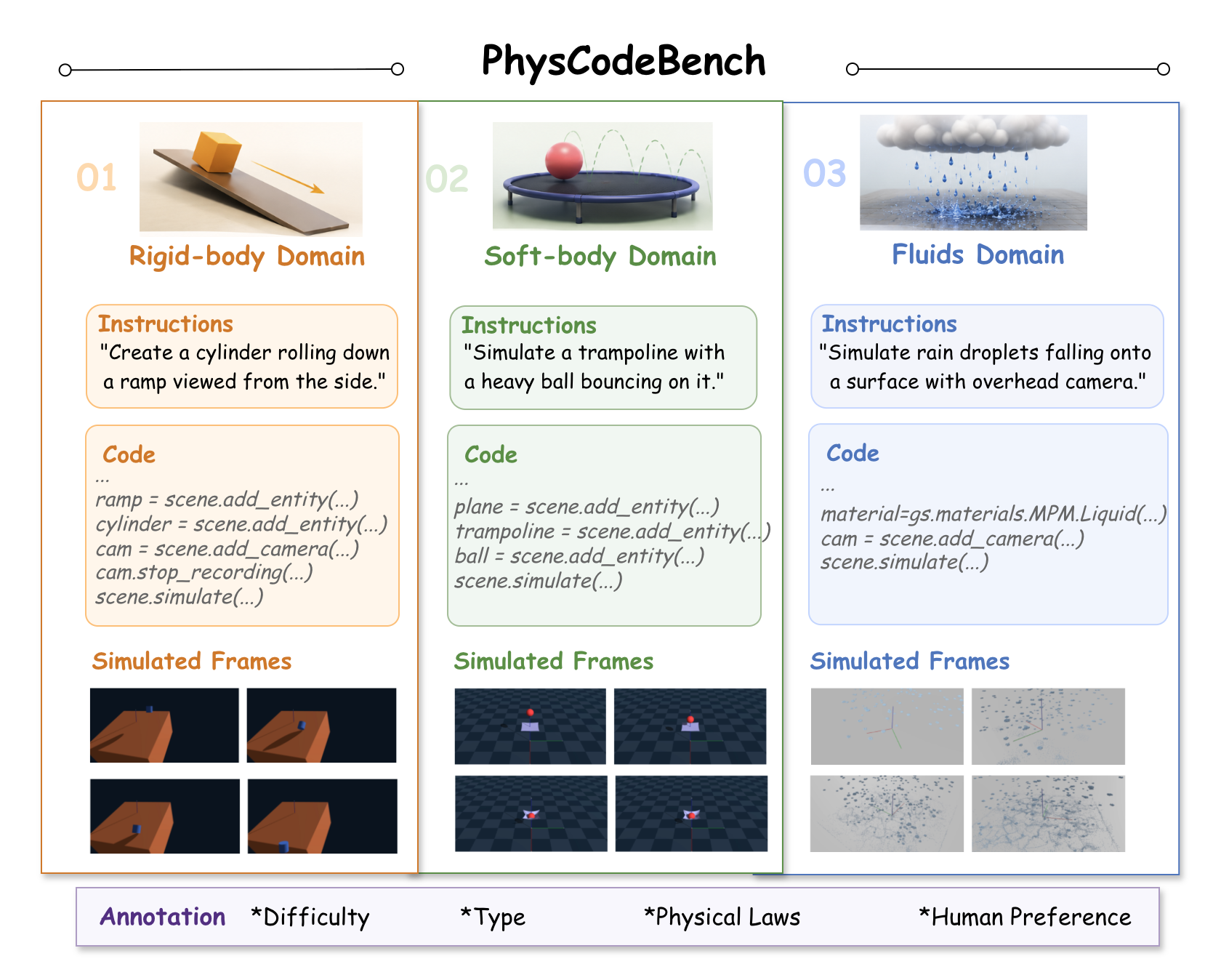}
    \caption{Representative PhysCodeBench examples from three physical domains. Each example pairs a natural-language instruction with an expert-validated Genesis implementation, rendered simulation frames, and annotation metadata (difficulty, scenario type, governing physical laws, human preference).}
    \label{fig:examples}
\end{figure}

Figure~\ref{fig:examples} shows representative examples. Table~\ref{tab:domain_dist} gives the domain-by-difficulty composition of all 1,200 examples. Table~\ref{tab:laws} shows the distribution of governing physical laws. Counts use the primary classification, and 376 examples (140 easy / 236 hard) additionally carry a secondary domain label (e.g., rigid bodies coupled with fluid).

The ``Other'' category covers less frequent principles such as surface tension, buoyancy, and material phase transitions. The test split (200 examples) contains 68 rigid-body, 52 soft-body, 46 fluid, and 34 mechanics examples (116 easy / 84 hard), mirroring the overall composition.

\subsection{Curation Funnel and Success Rates}
\label{app:funnel}

The full funnel is: 50 expert seed prompts (15 held out for the test split) $\rightarrow$ 2,000 LLM-expanded candidate instructions $\rightarrow$ 1,400 instructions after human filtering (clarity, physical well-posedness, diversity) $\rightarrow$ two-stage validation with expert modification, retaining \textbf{1,120 validated examples} (80.0\%, with 280 instructions discarded after three failed revision rounds). Of these 1,120 pool-derived examples, 1,000 (from training seeds) form the training set and 120 (from held-out seeds) enter the test set. The 15 held-out seeds were deliberately expanded at lower volume than the 35 training seeds, since their role is to fill a fixed-size test split rather than to maximize training data. Adding \textbf{80 expert-authored test instructions} (written, together with their reference implementations, entirely by experts without LLM involvement) yields the final \textbf{1,200 examples}. Only 430/1,400 (30.7\%) of pooled candidates pass both validation stages on the first attempt. Tables~\ref{tab:success} and \ref{tab:regen} report stage-wise rates and intervention causes.

Fluid dynamics and mechanics required the most revision rounds on average (2.4 and 2.1) versus rigid-body (1.3) and soft-body (1.8) scenarios.

The rates in Table~\ref{tab:success} are per instruction over the four generating models, counting an instruction as successful if any model's draft succeeds. They are therefore not comparable to the per-sample code scores of a single model in the main results table, which are lower.

\paragraph{Construction effort.} Curation was performed by five domain experts (three with physics backgrounds, two with computer-graphics backgrounds) over approximately 1,280 person-hours: 90 for seed authoring and candidate filtering, 700 for validation and code revision, 340 for metadata, preference, and assertion annotation, and 150 for the MuJoCo-native reference implementations and ported assertion sets. The dominant cost is validation, at roughly 35 minutes per retained example, since each revision round requires re-running the simulation and inspecting the rendered result.

\subsection{Generation Prompt Template}

All code generation (dataset curation and evaluation alike) uses the template below. Here \texttt{\{user\_prompt\}} is the instruction, \texttt{\{genesis\_documentation\}} the retrieved documentation context, and \texttt{\{relevant\_code\_examples\}} example programs from the official Genesis repository.

\begin{footnotesize}
\begin{verbatim}
You are an expert programmer specializing
in physical simulations using the Genesis
physics engine. Your task is to implement
the following physical scenario:

[INSTRUCTION]: {user_prompt}

Please generate Python code that implements
this scenario using the Genesis physics
engine. Your code should:
1. Initialize the Genesis environment with
   appropriate parameters
2. Create all necessary physical objects
   with realistic properties
3. Configure the correct physical
   interactions and constraints
4. Set up an appropriate camera angle to
   visualize the phenomenon
5. Run the simulation and save the output
   video

The code should be physically accurate,
following these laws:
- Respect conservation laws (energy,
  momentum, etc.)
- Use realistic parameters for mass,
  friction, elasticity, etc.
- Implement correct collision detection
  and response
- Apply appropriate forces and constraints

For the output specifications:
- Set the resolution to 1280x640 pixels
- Use a frame rate of 60 fps
- Generate a 5-second video
- Save the output as "genesis_video.mp4"
- Set visualization parameter to False

Here are relevant examples and
documentation to help you:
[CONTEXT]: {genesis_documentation}
[EXAMPLES]: {relevant_code_examples}
\end{verbatim}
\end{footnotesize}

\begin{table}[t]
\centering
\small
\begin{tabular}{lccc}
\toprule
\textbf{Domain} & \textbf{Easy} & \textbf{Hard} & \textbf{Total} \\
\midrule
Rigid-body physics & 246 & 164 & 410 \\
Soft-body physics  & 186 & 124 & 310 \\
Fluid dynamics     & 154 & 121 & 275 \\
Mechanics          & 123 &  82 & 205 \\
\midrule
\textbf{Total}     & 709 & 491 & 1,200 \\
\bottomrule
\end{tabular}
\caption{Domain and difficulty composition of PhysCodeBench (primary classification).}
\label{tab:domain_dist}
\end{table}

\begin{table}[t]
\centering
\small
\setlength{\tabcolsep}{4pt}
\begin{tabular}{lcccc}
\toprule
\textbf{Physical law} & \textbf{Rigid} & \textbf{Soft} & \textbf{Fluid} & \textbf{Mech.} \\
\midrule
Collisions      & 294 & 122 &  49 & 100 \\
Gravity         & 235 & 201 & 183 & 121 \\
Elasticity      & 145 & 225 &  31 &  75 \\
Friction        & 239 & 118 &  58 & 114 \\
Fluid dynamics  &  20 &  41 & 206 &  16 \\
Other           &  44 &  33 &  54 &  31 \\
\bottomrule
\end{tabular}
\caption{Occurrence of physical-law tags by domain (2,755 tags over 1,200 examples, mean 2.3 per example). Easy examples typically carry 1--2 tags and hard examples 3 or more, though difficulty also depends on the other criteria of Section~A.4.}
\label{tab:laws}
\end{table}

\begin{table}[t]
\centering
\small
\begin{tabular}{lc}
\toprule
\textbf{Stage (criterion)} & \textbf{Rate (\%)} \\
\midrule
First attempt: executes without error       & 64.2 \\
First attempt: passes both validation stages & 30.7 \\
After expert revision ($\leq$3 rounds): both stages & 80.0 \\
\bottomrule
\end{tabular}
\caption{Validation success rates over the 1,400-instruction pool. The first row uses the execution-only criterion. The latter two require both execution and simulation validation.}
\label{tab:success}
\end{table}

\begin{table}[t]
\centering
\small
\begin{tabular}{lc}
\toprule
\textbf{Issue type} & \textbf{Share (\%)} \\
\midrule
API usage errors                     & 42.8 \\
Physical parameter misconfiguration  & 28.6 \\
Syntax errors                        & 15.3 \\
Incompatible object interactions     &  9.2 \\
Other implementation issues          &  4.1 \\
\bottomrule
\end{tabular}
\caption{Issues requiring code modification or regeneration during curation.}
\label{tab:regen}
\end{table}

\subsection{Annotation Guidelines and Agreement}
\label{app:annotation}

\paragraph{Difficulty.} \emph{Easy}: 1--2 physical laws, common object interactions, standard parameter settings, single phase of matter. \emph{Hard}: 3+ physical laws, complex interactions, precise parameter tuning, multiple phases or state transitions. Three annotators label each example. Agreement is mean pairwise Cohen's $\kappa = 0.78$.

\paragraph{Physical-law tags.} Multi-label annotation from a fixed vocabulary. Agreement is mean pairwise Jaccard-based $\kappa = 0.71$.

\paragraph{Preference judgments.} For 600 scenarios we collect two independent validated implementations (600 implementation pairs) and ask five domain experts for pairwise preferences, yielding 1,980 preference judgments (3.3 raters per pair on average). Because the number of raters varies across pairs, we report Krippendorff's $\alpha = 0.68$ rather than Fleiss' $\kappa$, which assumes a fixed number of raters per item. Annotators weigh physical accuracy first, then code quality, then prompt adherence. The majority label per pair yields the 600 DPO training pairs for the Simulation Refiner.

\paragraph{Physical assertions.} During test-set annotation, experts author 1--3 executable assertions per instruction (436 total), each specifying a measurable property of the correct simulation with an explicit tolerance. Assertions are written from the instruction text alone, \emph{blind to any candidate or reference implementation}, and each assertion is reviewed by a second expert for entailment by the instruction. Assertions constraining parameters the instruction leaves unspecified are loosened or the constraint is added to the instruction. Where instructions specify object counts, attributes, or spatial composition, assertions check them from engine state (e.g., ``exactly four legs remain in contact with the ground''). As an oracle sanity check, the expert reference implementations pass 98.6\% of all assertions and 100\% of the execution and artifact checks, and score 44.1 of 50 on the perceptual axis. These four values are the oracle row of the main results table, and the perceptual score stays below 50 because CLIP similarity does not saturate even for a faithful rendering.

\section{PhysCodeEval Details}

\subsection{Code-Level Scoring}

$S_{\text{exec}}$ (25 pts): the program is executed in a sandboxed Docker environment with a 120\,s timeout, and any runtime error scores 0. Error categories (syntax, physics-parameter, API misuse, resource exhaustion) are logged for diagnosis but do not affect the binary score. $S_{\text{file}}$ (25 pts): checks existence of the required video artifact, minimum size (100\,KB), and format compliance (1280$\times$640, 60\,fps, 5\,s). Partial credit follows a fixed rubric weighting each requirement.

\subsection{Visual Scoring}
\label{app:visual}

$S_{\text{clip}}$ (25 pts): 10 evenly spaced frames are embedded with CLIP, and the mean cosine similarity $s$ to the instruction embedding is mapped to $[0,25]$ by $25\cdot\mathrm{clip}\!\left((s-s_{\min})/(s_{\max}-s_{\min}),0,1\right)$~\citep{hessel2021clipscore}. $S_{\text{motion}}$ (25 pts): a flow-based smoothness measure inspired by VBench's motion-quality principle~\citep{huang2024vbench}. The 60\,fps output video is temporally resampled to 25\,fps, identically for every system so that the resampling cannot bias the comparison, and RAFT optical flow~\citep{teed2020raft} is computed between consecutive frames. For each connected foreground region we take the $L_2$ norm of second-order velocity differences and average over regions and frames, so that scenes with more moving objects are not penalized, before mapping to $[0,25]$. Jitter, teleportation, and interpenetration inflate the second-order term and reduce the score. In pseudocode:

\begin{footnotesize}
\begin{verbatim}
frames = resample(video, fps=25)
flows = [RAFT(f[t], f[t+1]) for t]
j = []
for region in foreground(frames):
    v = mean_flow(region, flows)
    a = diff(v)          # acceleration
    jerk = diff(a)       # 2nd-order diff
    j.append(l2(jerk))
# lower mean jerk -> higher score
mj = mean(j)
S_motion = 25 * clip(1 - mj/J_max, 0, 1)
\end{verbatim}
\end{footnotesize}

The three calibration constants ($s_{\min}, s_{\max}, J_{\max}$) are fixed once for the whole benchmark, are never tuned per system or per scenario, and ship with the evaluation toolkit so that scores are reproducible.

\subsection{Direct Physical-Correctness Scoring}
\label{app:sphys}

All checks are computed from engine state queries (positions, velocities, angular velocities of every entity at every step), never from rendered video.

\paragraph{Conservation checks.} For scenario labels with well-defined invariants:
\begin{itemize}
    \item \emph{Momentum} (collision scenarios): $\lVert \Delta p \rVert / \lVert p_0 \rVert \le 0.05$ over the interaction window.
    \item \emph{Free fall}: least-squares acceleration fit within $5\%$ of $g$.
    \item \emph{Collision energy}: kinetic-energy retention $\ge 0.95$ for elastic collisions, and dissipation inside the scenario-labeled admissible interval for inelastic ones.
    \item \emph{Incline dynamics}: fitted acceleration within $5\%$ of $g(\sin\theta - \mu\cos\theta)$ for the labeled $\theta, \mu$.
\end{itemize}
For dissipative regimes (soft body, fluid) where global conservation is ill-defined, we instead verify necessary conditions: monotone energy dissipation (no spontaneous energy injection), terminal-state stability (bounded velocities at the horizon), and non-interpenetration (signed distances above a contact tolerance).

\paragraph{Assertion examples.}
\begin{itemize}
    \item ``Ball bouncing on trampoline'': first rebound height $\in [0.6h_0, 0.95h_0]$.
    \item ``Raindrops with ripples'': $\geq 3$ concentric surface waves within 0.3\,s of first impact.
    \item ``Cube on incline'': cube's displacement direction within $10^\circ$ of the downhill direction, with no penetration below the incline plane.
    \item ``Two-sphere elastic collision'': post-impact velocities match the analytic two-body solution within 5\%.
    \item ``Four-legged table drop'': exactly four legs in ground contact at rest, and the entity count matches the instruction.
\end{itemize}

\paragraph{Per-residual results.} Table~\ref{tab:conservation} breaks down the conservation subset (102 rigid-body and mechanics test cases) reported in the main paper.

\begin{table}[t]
\centering
\small
\setlength{\tabcolsep}{4pt}
\begin{tabular}{lccc}
\toprule
\textbf{Check} & \textbf{SMRF} & \textbf{Claude-3.5} & \textbf{\drdq{}+S+D} \\
\midrule
Momentum drift      & 83.7 & 37.5 & 44.8 \\
Free-fall / incline & 85.0 & 40.2 & 47.2 \\
Energy (elastic)    & 76.2 & 28.9 & 35.4 \\
Energy (inelastic)  & 80.7 & 34.6 & 41.4 \\
\midrule
\textbf{Aggregate}  & \textbf{81.4} & 35.3 & 42.2 \\
\bottomrule
\end{tabular}
\caption{Pass rates (\%) on trajectory-level conservation checks over the 102 rigid-body and mechanics test cases (5\% tolerance). Each check is evaluated only on the cases whose scenario label defines the corresponding invariant, so the row denominators differ. Aggregate is the unweighted mean of the four checks rather than a case-level rate.}
\label{tab:conservation}
\end{table}

\subsection{Cross-Engine Protocol}

For 120 of the 200 test instructions (covering all four domains, with fluid scenarios restricted to MuJoCo-supported particle approximations), experts authored MuJoCo-native reference implementations and ported the assertion sets. All PhysCodeEval axes are computed identically on MuJoCo: execution in a sandboxed MuJoCo environment, artifact checks, perceptual scores on rendered frames, and assertion checks from \texttt{mjData} state queries (on this subset, SMRF with the 250 adaptation samples passes 63.4\% of ported assertions versus 20.1\% for zero-shot Claude-3.5-Sonnet). Cross-engine retention percentages in the main paper use SMRF's Genesis score restricted to the same 120 instructions as the denominator. The 250 adaptation samples referenced in the main paper are API-mapping demonstrations (Genesis$\rightarrow$MuJoCo construct correspondences) containing no new physical scenarios. The EC fix-rate comparison (84\%$\rightarrow$79\%) is measured in the zero-shot-transfer condition.

\section{Training and Inference Details}
\label{app:training}

All agents are initialized from DeepSeek-R1-Distill-Qwen-32B and fine-tuned with LoRA (rank 64, applied to attention and MLP projections) using AdamW (weight decay 0.01), learning rate $10^{-5}$ with cosine schedule and 100 warmup steps, batch size 2 with gradient accumulation, 5 epochs for SFT and 3 for DPO ($\beta = 0.1$), early stopping on validation loss. Training used 2$\times$NVIDIA A100 80GB for $\approx$34 GPU-hours for the main configuration (cross-validation retrainings excluded). Random seed 42 is fixed for data ordering and initialization. Harvest and training scripts will be released.

At inference, long-context models (GPT-4o, Claude-3.5-Sonnet, Gemini-2.0-Pro, DeepSeek-R1) receive the full $\approx$100K-token Genesis documentation corpus (official examples, API reference, user guide). Models with 32K contexts (Qwen-2.5-32B, \drdq{}, QwQ-32B, and all SMRF agents) receive $\approx$20K tokens of BM25-retrieved~\citep{robertson2009bm25} sections keyed on the instruction's entities and physical laws, plus the top-3 most similar official examples. Generation: temperature 0.1, max 4,096 new tokens. Every test instruction is attempted 5 times and we report the mean (avg@5). The same protocol applies to all baselines and SMRF. The generation cap is a real constraint for the zero-shot reasoning-style open models (DeepSeek-R1, QwQ-32B, and \drdq{}), whose chains of thought can consume a large share of the budget before any code is emitted. It does not affect the comparison against Claude-3.5-Sonnet or against the fine-tuned single agent, both of which emit code directly, but we flag it in Section~F rather than claim it away.

\paragraph{Software environment.} Genesis \texttt{0.2.1}, MuJoCo \texttt{3.3.2}, PyTorch \texttt{2.6.0} with CUDA \texttt{12.4}, Python \texttt{3.10}, and the official RAFT reference implementation. Every engine version is pinned, because simulation results are version-sensitive. Evaluation runs inside a fixed Docker image whose full dependency manifest ships with the toolkit.

\paragraph{Context ablation.} To verify that the asymmetric documentation protocol neither handicaps nor favors either group, we ablate the context given to Claude-3.5-Sonnet: full 100K corpus 35.9, naive truncation to 30K 33.8, BM25-retrieved 20K 34.7. Retrieval recovers most of the full-context score, so the 32K-context models' protocol is not the driver of the SMRF--baseline gap (Table~\ref{tab:context}).

\begin{table}[t]
\centering
\small
\begin{tabular}{lc}
\toprule
\textbf{Context given to Claude-3.5-Sonnet} & \textbf{Total} \\
\midrule
Full documentation ($\approx$100K tokens) & 35.9 \\
Naive truncation (30K tokens)             & 33.8 \\
BM25-retrieved sections (20K tokens)      & 34.7 \\
\bottomrule
\end{tabular}
\caption{Context ablation for the documentation protocol.}
\label{tab:context}
\end{table}

\section{SMRF Implementation Details}

\subsection{Agent Loop}

Given instruction $x$, the SG produces $c_0$, which is executed in the engine. On failure with runtime feedback $e_t$, the EC produces $c_{t+1} = \text{EC}(x, c_t, e_t)$, for at most 3 rounds, after which the framework returns failure. On success, the SR produces the final program $\hat{c} = \text{SR}(x, c_t)$, adjusting physical parameters and code quality while preserving structure. Each agent is a separately fine-tuned LoRA specialization of the same backbone, so the framework's memory footprint is one backbone plus three adapters.

\subsection{Error-Corrector Tetrads}
\label{app:tetrads}

The EC trains on 520 tetrads $(x, c_{\text{bug}}, e, c_{\text{fix}})$ harvested from Stage-3 validation failures, restricted to the 1,000 training instructions (no tetrad derives from any test instruction): $x$ is the instruction, $c_{\text{bug}}$ the failing program, $e$ the runtime feedback (exception trace or expert-written physical diagnosis), and $c_{\text{fix}}$ the expert-corrected program. Loss is computed only on $c_{\text{fix}}$ tokens. Category distribution: 43\% API misuse, 28\% parameter miscalibration, 18\% boundary conditions, 7\% temporal discretization, 4\% other. This taxonomy groups tetrads by the physical or programmatic failure mode that the fix addresses, and is therefore not the same partition as the intervention-cause taxonomy of Table~\ref{tab:regen}, which counts why a curation example needed expert attention at all.

Three abbreviated examples:

\begin{footnotesize}
\begin{verbatim}
# Tetrad 1 (API misuse)
# e: TypeError: add_entity() got an
#    unexpected keyword 'friction_coef'
- surface = scene.add_entity(
-     gs.morphs.Plane(),
-     friction_coef=0.5)
+ surface = scene.add_entity(
+     gs.morphs.Plane(),
+     material=gs.materials.Rigid(
+         friction=0.5))

# Tetrad 2 (parameter miscalibration)
# e: expert diagnosis: restitution 1.4
#    injects energy; ball gains height
#    every bounce
- mat = gs.materials.Rigid(
-     restitution=1.4)
+ mat = gs.materials.Rigid(
+     restitution=0.85)

# Tetrad 3 (boundary condition)
# e: fluid particles escape the domain;
#    simulation diverges at t=1.8s
- scene = gs.Scene(
-     sim_options=gs.options.SimOptions())
+ scene = gs.Scene(
+     sim_options=gs.options.SimOptions(),
+     boundary=gs.options.Boundary(
+         lower=(-2,-2,0), upper=(2,2,4)))
\end{verbatim}
\end{footnotesize}

\subsection{Semi-Automated Benchmark Expansion}

To address the cost of expert curation, we fine-tune a prompt evaluator (accept/reject with rationale) on the Stage-1 filtering decisions over the 2,000 candidate instructions. In a held-out trial on 200 fresh candidates, routing candidates through the evaluator before expert review reduced expert filtering time by 64\% at equal acceptance quality. The released toolkit packages this evaluator with the validation harness, so community contributions require one expert pass rather than full pipeline review.

\section{Additional Results}

\subsection{Performance by Physical Domain}

\begin{table}[t]
\centering
\small
\setlength{\tabcolsep}{3.5pt}
\begin{tabular}{lcccc}
\toprule
\textbf{Model} & \textbf{Rigid} & \textbf{Soft} & \textbf{Fluid} & \textbf{Mech.} \\
\midrule
GPT-4o              & 37.6 & 30.8 & 27.9 & 38.8 \\
Claude-3.5-Sonnet   & 39.9 & 32.9 & 29.9 & 40.8 \\
Gemini-2.0-Pro      & 35.5 & 28.5 & 25.6 & 36.5 \\
DeepSeek-R1         & 33.4 & 26.5 & 23.5 & 34.4 \\
\drdq{}             & 31.5 & 24.6 & 21.7 & 32.5 \\
Qwen-2.5-32B        &  2.7 &  1.2 &  0.7 &  2.9 \\
QwQ-32B             & 18.3 & 13.4 & 11.4 & 18.9 \\
\drdq{} + SFT       & 39.3 & 32.4 & 29.5 & 40.3 \\
\drdq{} + SFT + DPO & 41.4 & 34.5 & 31.5 & 42.4 \\
SMRF (base)         & 37.6 & 30.7 & 27.7 & 38.5 \\
SMRF + SFT          & 66.2 & 58.4 & 54.3 & 67.1 \\
SMRF + SFT + DPO    & \textbf{72.3} & \textbf{63.2} & \textbf{59.4} & \textbf{73.6} \\
\bottomrule
\end{tabular}
\caption{Total score by physical domain on the test set.}
\label{tab:domain_scores}
\end{table}

Table~\ref{tab:domain_scores} shows that all systems find fluid dynamics hardest (multi-phase behavior, stability constraints) and mechanics/rigid-body easiest. SMRF leads the best proprietary baseline (Claude-3.5-Sonnet) by 29.5--32.8 points and the strongest fine-tuned single agent (\drdq{}+SFT+DPO) by 27.9--31.2 points in every domain, so the advantage is not an artifact of one scenario family.

\subsection{Error Correction Analysis}

\begin{table}[t]
\centering
\small
\begin{tabular}{lcc}
\toprule
\textbf{Error type} & \textbf{Freq.\ (\%)} & \textbf{Fixed (\%)} \\
\midrule
API usage errors            & 41 & 88 \\
Parameter miscalibration    & 30 & 84 \\
Boundary condition errors   & 17 & 79 \\
Temporal discretization     &  8 & 75 \\
Other                       &  4 & 71 \\
\bottomrule
\end{tabular}
\caption{Errors identified by the EC during SMRF inference on the test set and correction success rates. The frequency distribution is measured at inference time and differs from the training-tetrad distribution in Section~D.2, which is harvested from curation-stage failures of frontier LLMs.}
\label{tab:error_types}
\end{table}

Typical EC diagnoses include mismatched API parameter names, camera configurations that miss the phenomenon, and physically inadmissible values (``restitution 1.4 injects energy'' or ``gravity 50\,m/s$^2$ unrealistic for this scenario''). Frequencies and success rates by category are in Table~\ref{tab:error_types}.

\subsection{Single-Agent Iterative Refinement and Best-of-$k$}

\begin{table}[t]
\centering
\small
\begin{tabular}{lc}
\toprule
\textbf{Approach} & \textbf{Total} \\
\midrule
GPT-4o (3 corrections)              & 38.1 \\
Claude-3.5-Sonnet (3 corrections)   & 39.8 \\
Claude-3.5-Sonnet (best-of-5, CLIP-selected) & 40.4 \\
\drdq{} + SFT + DPO (3 corrections) & 41.0 \\
SMRF + SFT + DPO                    & \textbf{67.2} \\
\midrule
Improvement over strongest alternative & \textbf{+26.2} \\
\bottomrule
\end{tabular}
\caption{Single-agent self-refinement (3 correction rounds, same budget as SMRF) and CLIP-selected best-of-5 sampling vs.\ SMRF.}
\label{tab:single_agent}
\end{table}

With an identical correction budget, the best single-agent self-refinement reaches 41.0, and boosting Claude-3.5-Sonnet with best-of-5 sampling (selecting by CLIPScore) reaches only 40.4 at 5$\times$ the generation cost (Table~\ref{tab:single_agent}). The gap to SMRF (+26.2 over the strongest alternative) isolates the value of \emph{specialized} correction and refinement from mere iteration or sampling.

\subsection{VLM-Based Physical Plausibility}

\begin{table}[t]
\centering
\small
\setlength{\tabcolsep}{4pt}
\begin{tabular}{lcc|cc}
\toprule
 & \multicolumn{2}{c|}{\textbf{GPT-4o judge}} & \multicolumn{2}{c}{\textbf{Gemini judge}} \\
\textbf{Aspect} & \textbf{SMRF} & \textbf{Cl-3.5} & \textbf{SMRF} & \textbf{Cl-3.5} \\
\midrule
Gravity              & 4.6 & 3.8 & 4.5 & 3.7 \\
Collision dynamics   & 4.4 & 3.5 & 4.2 & 3.5 \\
Fluid behavior       & 4.1 & 3.2 & 4.1 & 3.1 \\
Object motion        & 4.5 & 3.7 & 4.3 & 3.5 \\
Temporal consistency & 4.2 & 3.6 & 4.1 & 3.5 \\
\midrule
\textbf{Overall (mean)} & \textbf{4.4} & 3.6 & \textbf{4.2} & 3.5 \\
\bottomrule
\end{tabular}
\caption{VLM-rated physical plausibility (1--5) on 100 sampled test videos, 3 runs per sample (std $<$ 0.3), under two judges. Overall = mean of the five aspects.}
\label{tab:vlm}
\end{table}

Because GPT-4o is also an evaluated baseline, we cross-check with a second judge from a different model family (Gemini-2.0-Pro). Rankings and margins agree across judges (Table~\ref{tab:vlm}). We note that both judges are themselves evaluated baselines, so these ratings are supplementary evidence only: the non-circular physical check is the state-based measurement of Section~B.3, which involves no model judge.

\subsection{User-Study Protocol}

The ten participants (six graduate students, four professional developers, all with physics-simulation experience) were recruited from the authors' institutions, participated voluntarily with informed consent, and were compensated at standard local hourly rates. Each rated anonymized, randomly ordered implementations, and raters were blind to which system produced which output. Every participant rated all 200 implementations (40 prompts $\times$ 5 systems) over four sessions of at most 90 minutes, so the design is fully crossed and ICC(2,1) applies. The metric-validity correlation in the main paper uses the best proprietary baseline, the strongest single agent, and SMRF.

\subsection{Case Studies}

\paragraph{Rigid body.} For an unequal-mass elastic collision, the SG's draft used incorrect collision-response coefficients. The EC flagged the API misuse and restored the analytic post-impact velocities, and the SR tuned restitution so the kinetic-energy retention check passed at 0.97.

\paragraph{Fluid.} For a rain-droplet scenario, the draft's surface-tension setting produced droplet merging without splashes. The EC identified the parameter miscalibration, and the SR's preference-aligned refinement produced ripple propagation satisfying the ``$\geq$3 ripples in 0.3\,s'' assertion.

\paragraph{Soft body.} For cloth draping over a sphere, the draft's stiffness caused unbounded stretching. The EC corrected the elasticity range, and the terminal-state stability check then passed with bounded strain.

\subsection{Additional Qualitative Comparisons}

\begin{figure}[t]
    \centering
    \includegraphics[width=0.9\textwidth]{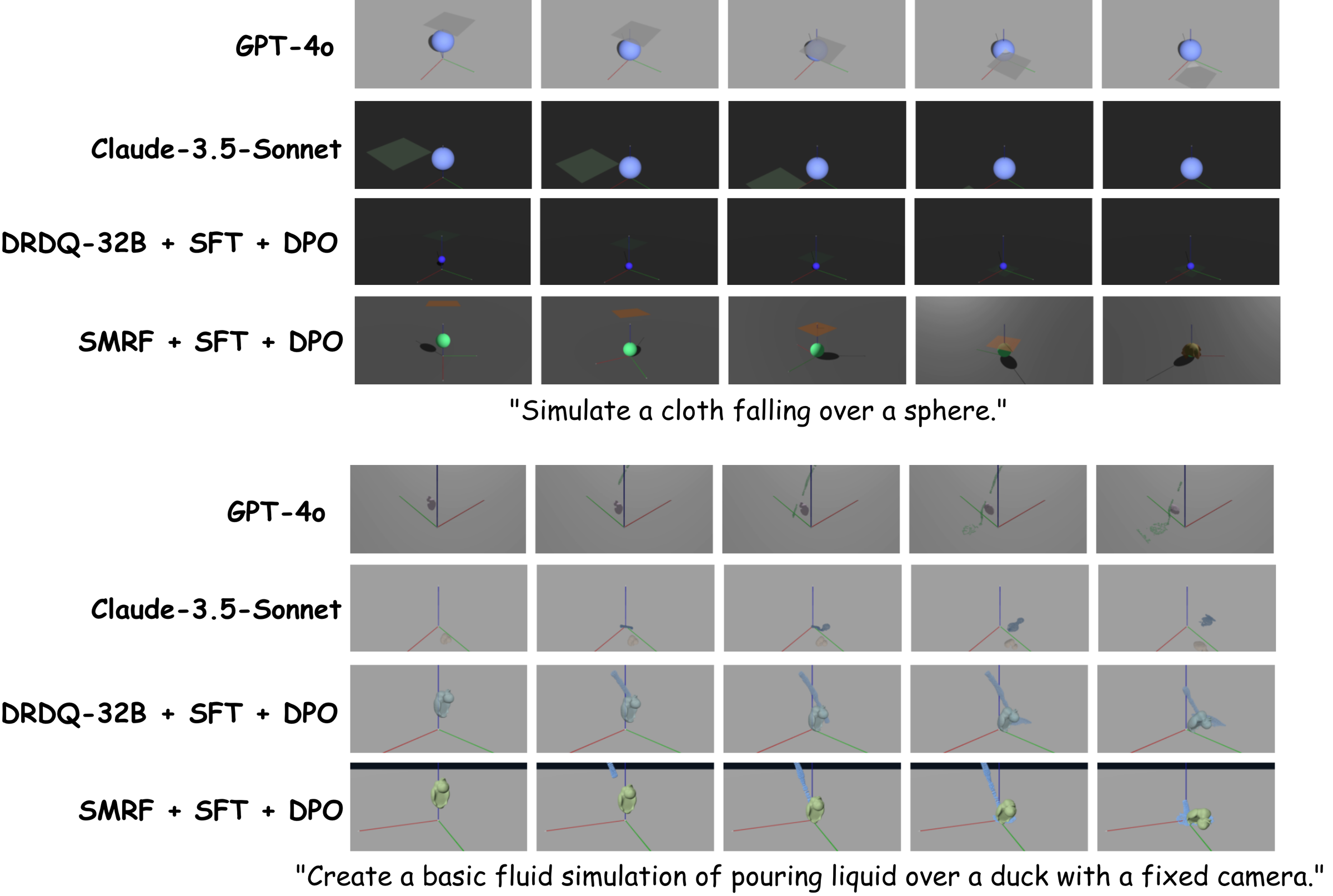}
    \caption{Additional qualitative comparisons. Top (``cloth falling over a sphere''): baselines leave the sphere bare (cloth absent or rigid), while SMRF drapes the cloth with contact-consistent folds. Bottom (``pouring liquid over a duck''): baselines emit sparse disconnected particles, while SMRF produces a continuous stream interacting with the surface.}
    \label{fig:appendix_qual}
\end{figure}

Figure~\ref{fig:appendix_qual} shows two further test instructions with frame sequences for all four systems.

\subsection{Contamination Audit Details}

Semantic: each of the 200 test instructions is embedded (text-embedding-3-large) and compared to all 1,000 training instructions. Nearest-neighbor cosine similarity has mean 0.41 and max 0.63, with zero pairs above the 0.85 near-duplicate threshold. Lexical: BM25 retrieval of the closest training instruction gives max token-level Jaccard 0.27. Code: five zero-shot samples per test prompt from each of the strongest baselines never exceed CodeBLEU~\citep{ren2020codebleu} 0.7 against the corresponding test reference implementations, and expert rewriting during curation leaves mean CodeBLEU 0.46 between final references and original LLM drafts. Prompt-style control: on the hard subset, SMRF scores 52.1 on held-out-seed expansions vs.\ 53.9 on expert-authored instructions (1.8-point gap), so performance does not depend on the LLM-expansion style shared with training data. Statistical robustness: 1,000-resample bootstrap 95\% CIs by domain do not overlap between SMRF and any baseline, and seed-grouped 5-fold cross-validation over the 1,120 pool-derived examples gives a total-score std of 1.4.

\section{Limitations}

PhysCodeBench targets symbolic simulation through high-level engine APIs, not the implementation of solvers or numerical methods. Mastering the former is a prerequisite for the latter, but our scores say nothing about a model's ability to write, e.g., a stable MPM solver. Domain coverage excludes electromagnetism, thermodynamics beyond simple dissipation, and relativistic or quantum effects. The dissipative-scenario checks are necessary conditions rather than sufficient ones: a simulation can pass them while still deviating from the described phenomenon in ways the assertions do not capture. We also report conservation residuals as an aggregate only for the 102 rigid-body and mechanics cases. For the dissipative regimes the necessary conditions are checked per scenario and surface through the assertion set, but we do not summarize them as a standalone rate, since their admissible bands are scenario-specific and not comparable across scenarios. The Error Corrector is triggered only by execution failure, so semantically incorrect but crash-free programs bypass correction at inference time. The state-based checks detect (but do not yet repair) such cases, and closing this loop is future work. Cross-engine evaluation currently covers MuJoCo for 120 of 200 test cases, and fluid coverage there is approximate.

Two limits concern coverage of our own metrics. Executable assertions exist for the 200 test examples (436 assertions) and conservation residuals for the 102 rigid-body and mechanics test cases, so \sphys{} and the conservation diagnostic characterize the evaluation split rather than all 1,200 examples. Extending assertion coverage to the training split is the main outstanding curation cost.

A protocol caveat also deserves stating. All systems share a 4,096-token generation budget, which binds hardest on the zero-shot reasoning-style open models (DeepSeek-R1, QwQ-32B, and \drdq{}), whose chains of thought precede any code, and correspondingly on SMRF (base), which runs unspecialized \drdq{} agents. Those rows should be read as scores under a fixed budget rather than as ceilings for those models. The two headline comparisons are not affected: Claude-3.5-Sonnet is not a long-chain reasoning model and comfortably fits the budget, and the strongest single agent (\drdq{}+SFT+DPO) is fine-tuned on the same reference implementations as the SMRF agents, so both sides of that comparison use the budget the same way.

Finally, SMRF triples inference cost relative to a single agent, which matters for interactive use.

\section{LLM Usage Statement}

LLMs were used in three declared roles. (1) \emph{Dataset construction}: GPT-4o, Claude-3.5-Sonnet, Gemini-2.0-Pro, and GitHub Copilot expanded the 50 expert seed prompts into 2,000 candidates and drafted candidate implementations. Every retained example passed two-stage validation and expert rewriting (Section~A.2), and 80 test instructions were authored entirely without LLMs. (2) \emph{Evaluation subjects}: three of the four construction models (GPT-4o, Claude-3.5-Sonnet, Gemini-2.0-Pro) are also evaluated as baselines, alongside open-source models. (3) \emph{Writing}: language polishing only. All technical content, experiments, and analyses are by the authors.

%% file: physcodebench.bib
@inproceedings{code_as_policies,
  author={Liang, Jacky and Huang, Wenlong and Xia, Fei and Xu, Peng and Hausman, Karol and Ichter, Brian and Florence, Pete and Zeng, Andy},
  booktitle={2023 IEEE International Conference on Robotics and Automation (ICRA)},
  title={Code as Policies: Language Model Programs for Embodied Control},
  year={2023},
  pages={9493--9500},
  doi={10.1109/ICRA48891.2023.10160591}
}

@misc{huang2023voxposercomposable3dvalue,
      title={VoxPoser: Composable 3D Value Maps for Robotic Manipulation with Language Models}, 
      author={Wenlong Huang and Chen Wang and Ruohan Zhang and Yunzhu Li and Jiajun Wu and Li Fei-Fei},
      year={2023},
      eprint={2307.05973},
      archivePrefix={arXiv},
      primaryClass={cs.RO},
      url={https://arxiv.org/abs/2307.05973}, 
}

@inproceedings{simgen,
 author = {Zhou, Yunsong and Simon, Michael and Peng, Zhenghao and Mo, Sicheng and Zhu, Hongzi and Guo, Minyi and Zhou, Bolei},
 booktitle = {Advances in Neural Information Processing Systems},
 editor = {A. Globerson and L. Mackey and D. Belgrave and A. Fan and U. Paquet and J. Tomczak and C. Zhang},
 pages = {48838--48874},
 publisher = {Curran Associates, Inc.},
 title = {SimGen: Simulator-conditioned Driving Scene Generation},
 url = {https://proceedings.neurips.cc/paper_files/paper/2024/file/57c5a7c83b056d74bc97b7db36bd3649-Paper-Conference.pdf},
 volume = {37},
 year = {2024}
}

@INPROCEEDINGS{mujoco,
  author={Todorov, Emanuel and Erez, Tom and Tassa, Yuval},
  booktitle={2012 IEEE/RSJ International Conference on Intelligent Robots and Systems}, 
  title={MuJoCo: A physics engine for model-based control}, 
  year={2012},
  volume={},
  number={},
  pages={5026-5033},
  doi={10.1109/IROS.2012.6386109}}

@article{liu2022mind,
  title={Mind's {Eye}: Grounded language model reasoning through simulation},
  author={Liu, Ruibo and Wei, Jason and Gu, Shixiang Shane and Wu, Te-Yen and Vosoughi, Soroush and Cui, Claire and Zhou, Denny and Dai, Andrew M},
  journal={arXiv preprint arXiv:2210.05359},
  year={2022}
}

@article{rafailov2023direct,
  title={Direct preference optimization: Your language model is secretly a reward model},
  author={Rafailov, Rafael and Sharma, Archit and Mitchell, Eric and Manning, Christopher D and Ermon, Stefano and Finn, Chelsea},
  journal={Advances in Neural Information Processing Systems},
  volume={36},
  pages={53728--53741},
  year={2023}
}

@article{ouyang2022training,
  title={Training language models to follow instructions with human feedback},
  author={Ouyang, Long and Wu, Jeffrey and Jiang, Xu and Almeida, Diogo and Wainwright, Carroll and Mishkin, Pamela and Zhang, Chong and Agarwal, Sandhini and Slama, Katarina and Ray, Alex and others},
  journal={Advances in Neural Information Processing Systems},
  volume={35},
  pages={27730--27744},
  year={2022}
}

@article{le2023codechain,
  title={{CodeChain}: Towards modular code generation through chain of self-revisions with representative sub-modules},
  author={Le, Hung and Chen, Hailin and Saha, Amrita and Gokul, Akash and Sahoo, Doyen and Joty, Shafiq},
  journal={arXiv preprint arXiv:2310.08992},
  year={2023}
}

@article{chen2023teaching,
  title={Teaching large language models to self-debug},
  author={Chen, Xinyun and Lin, Maxwell and Sch{\"a}rli, Nathanael and Zhou, Denny},
  journal={arXiv preprint arXiv:2304.05128},
  year={2023}
}

@article{qian2023chatdev,
  title={{ChatDev}: Communicative agents for software development},
  author={Qian, Chen and Liu, Wei and Liu, Hongzhang and Chen, Nuo and Dang, Yufan and Li, Jiahao and Yang, Cheng and Chen, Weize and Su, Yusheng and Cong, Xin and others},
  journal={arXiv preprint arXiv:2307.07924},
  year={2023}
}

@article{wu2023autogen,
  title={{AutoGen}: Enabling next-gen {LLM} applications via multi-agent conversation},
  author={Wu, Qingyun and Bansal, Gagan and Zhang, Jieyu and Wu, Yiran and Li, Beibin and Zhu, Erkang and Jiang, Li and Zhang, Xiaoyun and Zhang, Shaokun and Liu, Jiale and others},
  journal={arXiv preprint arXiv:2308.08155},
  year={2023}
}

@article{luo2023wizardcoder,
  title={{WizardCoder}: Empowering code large language models with {Evol-Instruct}},
  author={Luo, Ziyang and Xu, Can and Zhao, Pu and Sun, Qingfeng and Geng, Xiubo and Hu, Wenxiang and Tao, Chongyang and Ma, Jing and Lin, Qingwei and Jiang, Daxin},
  journal={arXiv preprint arXiv:2306.08568},
  year={2023}
}

@article{austin2021program,
  title={Program synthesis with large language models},
  author={Austin, Jacob and Odena, Augustus and Nye, Maxwell and Bosma, Maarten and Michalewski, Henryk and Dohan, David and Jiang, Ellen and Cai, Carrie and Terry, Michael and Le, Quoc and others},
  journal={arXiv preprint arXiv:2108.07732},
  year={2021}
}

@article{roziere2023code,
  title={Code {Llama}: Open foundation models for code},
  author={Roziere, Baptiste and Gehring, Jonas and Gloeckle, Fabian and Sootla, Sten and Gat, Itai and Tan, Xiaoqing Ellen and Adi, Yossi and Liu, Jingyu and Sauvestre, Romain and Remez, Tal and others},
  journal={arXiv preprint arXiv:2308.12950},
  year={2023}
}

@article{takamoto2022pdebench,
  title={{PDEBench}: An extensive benchmark for scientific machine learning},
  author={Takamoto, Makoto and Praditia, Timothy and Leiteritz, Raphael and MacKinlay, Daniel and Alesiani, Francesco and Pfl{\"u}ger, Dirk and Niepert, Mathias},
  journal={Advances in Neural Information Processing Systems},
  volume={35},
  pages={1596--1611},
  year={2022}
}

@article{cherian2024llmphy,
  title={{LLMPhy}: Parameter-identifiable physical reasoning combining large language models and physics engines},
  author={Cherian, Anoop and Corcodel, Radu and Jain, Siddarth and Romeres, Diego},
  journal={arXiv preprint arXiv:2411.08027},
  year={2024}
}

@article{ren2020codebleu,
  title={{CodeBLEU}: A method for automatic evaluation of code synthesis},
  author={Ren, Shuo and Guo, Daya and Lu, Shuai and Zhou, Long and Liu, Shujie and Tang, Duyu and Sundaresan, Neel and Zhou, Ming and Blanco, Ambrosio and Ma, Shuai},
  journal={arXiv preprint arXiv:2009.10297},
  year={2020}
}

@inproceedings{teed2020raft,
  title={{RAFT}: Recurrent all-pairs field transforms for optical flow},
  author={Teed, Zachary and Deng, Jia},
  booktitle={Computer Vision -- ECCV 2020},
  pages={402--419},
  year={2020},
  publisher={Springer}
}

@article{robertson2009bm25,
  title={The probabilistic relevance framework: {BM25} and beyond},
  author={Robertson, Stephen and Zaragoza, Hugo},
  journal={Foundations and Trends in Information Retrieval},
  volume={3},
  number={4},
  pages={333--389},
  year={2009},
  publisher={Now Publishers},
  doi={10.1561/1500000019}
}

@inproceedings{wang2024gensim,
  title={{GenSim}: Generating robotic simulation tasks via large language models},
  author={Wang, Lirui and Ling, Yiyang and Yuan, Zhecheng and Shridhar, Mohit and Bao, Chen and Qin, Yuzhe and Wang, Bailin and Xu, Huazhe and Wang, Xiaolong},
  booktitle={International Conference on Learning Representations},
  year={2024}
}

@inproceedings{wang2024robogen,
  title={{RoboGen}: Towards unleashing infinite data for automated robot learning via generative simulation},
  author={Wang, Yufei and Xian, Zhou and Chen, Feng and Wang, Tsun-Hsuan and Wang, Yian and Fragkiadaki, Katerina and Erickson, Zackory and Held, David and Gan, Chuang},
  booktitle={Proceedings of the 41st International Conference on Machine Learning},
  pages={51936--51983},
  volume={235},
  series={Proceedings of Machine Learning Research},
  year={2024},
  publisher={PMLR}
}

@inproceedings{hu2024scenecraft,
  title={{SceneCraft}: An {LLM} agent for synthesizing 3D scenes as {Blender} code},
  author={Hu, Ziniu and Iscen, Ahmet and Jain, Aashi and Kipf, Thomas and Yue, Yisong and Ross, David A and Schmid, Cordelia and Fathi, Alireza},
  booktitle={Proceedings of the 41st International Conference on Machine Learning},
  pages={19252--19282},
  volume={235},
  series={Proceedings of Machine Learning Research},
  year={2024},
  publisher={PMLR}
}

@inproceedings{ma2024eureka,
  title={{Eureka}: Human-level reward design via coding large language models},
  author={Ma, Yecheng Jason and Liang, William and Wang, Guanzhi and Huang, De-An and Bastani, Osbert and Jayaraman, Dinesh and Zhu, Yuke and Fan, Linxi and Anandkumar, Anima},
  booktitle={International Conference on Learning Representations},
  year={2024}
}

@article{zheng2024contphy,
  title={{ContPhy}: Continuum physical concept learning and reasoning from videos},
  author={Zheng, Zhicheng and Yan, Xin and Chen, Zhenfang and Wang, Jingzhou and Lim, Qin Zhi Eddie and Tenenbaum, Joshua B and Gan, Chuang},
  journal={arXiv preprint arXiv:2402.06119},
  year={2024}
}

@article{yi2019clevrer,
  title={{CLEVRER}: Collision events for video representation and reasoning},
  author={Yi, Kexin and Gan, Chuang and Li, Yunzhu and Kohli, Pushmeet and Wu, Jiajun and Torralba, Antonio and Tenenbaum, Joshua B},
  journal={arXiv preprint arXiv:1910.01442},
  year={2019}
}

@article{hessel2021clipscore,
  title={{CLIPScore}: A reference-free evaluation metric for image captioning},
  author={Hessel, Jack and Holtzman, Ari and Forbes, Maxwell and Bras, Ronan Le and Choi, Yejin},
  journal={arXiv preprint arXiv:2104.08718},
  year={2021}
}

@inproceedings{huang2024vbench,
  title={{VBench}: Comprehensive benchmark suite for video generative models},
  author={Huang, Ziqi and He, Yinan and Yu, Jiashuo and Zhang, Fan and Si, Chenyang and Jiang, Yuming and Zhang, Yuanhan and Wu, Tianxing and Jin, Qingyang and Chanpaisit, Nattapol and others},
  booktitle={Proceedings of the IEEE/CVF Conference on Computer Vision and Pattern Recognition},
  pages={21807--21818},
  year={2024}
}

@article{tung2023physion++,
  title={Physion++: Evaluating physical scene understanding that requires online inference of different physical properties},
  author={Tung, Hsiao-Yu and Ding, Mingyu and Chen, Zhenfang and Bear, Daniel and Gan, Chuang and Tenenbaum, Josh and Yamins, Dan and Fan, Judith and Smith, Kevin},
  journal={Advances in Neural Information Processing Systems},
  volume={36},
  pages={67048--67068},
  year={2023}
}

@article{bear2021physion,
  title={Physion: Evaluating physical prediction from vision in humans and machines},
  author={Bear, Daniel M and Wang, Elias and Mrowca, Damian and Binder, Felix J and Tung, Hsiao-Yu Fish and Pramod, RT and Holdaway, Cameron and Tao, Sirui and Smith, Kevin and Sun, Fan-Yun and others},
  journal={arXiv preprint arXiv:2106.08261},
  year={2021}
}

@article{chen2021evaluating,
  title={Evaluating large language models trained on code},
  author={Chen, Mark and Tworek, Jerry and Jun, Heewoo and Yuan, Qiming and Pinto, Henrique Ponde De Oliveira and Kaplan, Jared and Edwards, Harri and Burda, Yuri and Joseph, Nicholas and Brockman, Greg and others},
  journal={arXiv preprint arXiv:2107.03374},
  year={2021}
}

@misc{Genesis,
  author = {{Genesis Authors}},
  title = {Genesis: A Universal and Generative Physics Engine for Robotics and Beyond},
  month = {December},
  year = {2024},
  howpublished = {\url{https://github.com/Genesis-Embodied-AI/Genesis}},
  note = {Accessed: 2026-07-01},
}

@misc{drdq_32b,
    author = {DeepSeek-AI},
    title = {DeepSeek-R1-Distill-Qwen-32B},
    howpublished = {\url{https://huggingface.co/deepseek-ai/DeepSeek-R1-Distill-Qwen-32B}},
    note = {Accessed: 2026-07-01},
    year = {2025}
}

@misc{gemini_2_pro,
    author = {{Google DeepMind}},
    title = {Gemini 2.0 Model Updates: 2.0 Flash, Flash-Lite, Pro Experimental},
    howpublished = {\url{https://blog.google/innovation-and-ai/models-and-research/google-deepmind/gemini-model-updates-february-2025/}},
    note = {Accessed: 2026-07-01},
    year = {2025}
}

@misc{github_copilot,
    author = {{GitHub}},
    title = {GitHub Copilot},
    howpublished = {\url{https://github.com/features/copilot}},
    note = {Accessed: 2026-07-01},
    year = {2025}
}

@misc{claude_35,
    author = {{Anthropic}},
    title = {Claude 3.5 Sonnet},
    howpublished = {\url{https://www.anthropic.com/news/claude-3-5-sonnet}},
    note = {Accessed: 2026-07-01},
    year = {2024}
}

@misc{qwq,
    author = {{Qwen Team}},
    title = {QwQ-32B: Embracing the Power of Reinforcement Learning},
    howpublished = {\url{https://qwenlm.github.io/blog/qwq-32b/}},
    note = {Accessed: 2026-07-01},
    year = {2025}
}

@article{hurst2024gpt,
  title={{GPT-4o} system card},
  author={Hurst, Aaron and Lerer, Adam and Goucher, Adam P and Perelman, Adam and Ramesh, Aditya and Clark, Aidan and Ostrow, AJ and Welihinda, Akila and Hayes, Alan and Radford, Alec and others},
  journal={arXiv preprint arXiv:2410.21276},
  year={2024}
}

@article{guo2025deepseek,
  title={{DeepSeek-R1}: Incentivizing reasoning capability in {LLMs} via reinforcement learning},
  author={Guo, Daya and Yang, Dejian and Zhang, Haowei and Song, Junxiao and Zhang, Ruoyu and Xu, Runxin and Zhu, Qihao and Ma, Shirong and Wang, Peiyi and Bi, Xiao and others},
  journal={arXiv preprint arXiv:2501.12948},
  year={2025}
}

@article{yang2024qwen2,
  title={{Qwen2.5} technical report},
  author={Yang, An and Yang, Baosong and Zhang, Beichen and Hui, Binyuan and Zheng, Bo and Yu, Bowen and Li, Chengyuan and Liu, Dayiheng and Huang, Fei and Wei, Haoran and others},
  journal={arXiv preprint arXiv:2412.15115},
  year={2024}
}

@article{qiu2025phybench,
  title={PHYBench: Holistic Evaluation of Physical Perception and Reasoning in Large Language Models},
  author={Qiu, Shi and Guo, Shaoyang and Song, Zhuo-Yang and Sun, Yunbo and Cai, Zeyu and Wei, Jiashen and Luo, Tianyu and Yin, Yixuan and Zhang, Haoxu and Hu, Yi and others},
  journal={arXiv preprint arXiv:2504.16074},
  year={2025}
}
